\documentclass{article} % For LaTeX2e
\usepackage{iclr2025,times}

\usepackage{amsmath,amsfonts,bm}

\def\eqref#1{equation~\ref{#1}}
\def\1{\bm{1}}

\def\va{{\bm{a}}}

\def\vh{{\bm{h}}}

\def\vk{{\bm{k}}}

\def\vo{{\bm{o}}}

\def\vq{{\bm{q}}}
\def\vr{{\bm{r}}}

\def\vv{{\bm{v}}}

\def\vx{{\bm{x}}}
\def\vy{{\bm{y}}}

\def\mD{{\bm{D}}}

\def\mK{{\bm{K}}}

\def\mQ{{\bm{Q}}}

\def\mV{{\bm{V}}}
\def\mW{{\bm{W}}}
\def\mX{{\bm{X}}}

\DeclareMathAlphabet{\mathsfit}{\encodingdefault}{\sfdefault}{m}{sl}
\SetMathAlphabet{\mathsfit}{bold}{\encodingdefault}{\sfdefault}{bx}{n}

\def\sI{{\mathbb{I}}}

\def\sR{{\mathbb{R}}}

\newcommand{\R}{\mathbb{R}}

\usepackage[hidelinks]{hyperref}
\usepackage{url}
\usepackage{graphicx}
\usepackage{caption}
\usepackage{subcaption}
\usepackage{booktabs} % for professional tables
\usepackage{lipsum}
\usepackage{xspace}

\usepackage{glossaries}
\glsdisablehyper
\usepackage{algorithm}
\usepackage{algpseudocode}
\usepackage{multirow}
\usepackage{makecell}
\usepackage{wrapfig}
\usepackage{subcaption}
\usepackage{color,xcolor,framed}
\definecolor{lightgray}{gray}{0.9}
\definecolor{lavender}{RGB}{230,230,250}

\newcommand{\algoname}{{\fontfamily{bch}\selectfont Lexico}\xspace}

\newcommand{\new}[1]{{\color{black} #1}}

\newacronym{llm}{LLM}{Large Language Model}
\newacronym{csr}{CSR}{compressed sparse row}
\newacronym{omp}{OMP}{Orthogonal Matching Pursuit}
\newacronym{kv}{KV}{key-value}
\newacronym{gqa}{GQA}{Grouped Query Attention}

\title{\algoname:  Extreme KV Cache Compression via Sparse Coding over Universal Dictionaries}

\author{
\centerline{
Junhyuck Kim$^{k}$, Jongho Park$^{k}$, Jaewoong Cho$^{k}$, Dimitris Papailiopoulos$^{w,ms}$} \\
\centerline{ $^{k}$KRAFTON, $^{w}$University of Wisconsin-Madison, $^{ms}$Microsoft Research }
}

\iclrfinalcopy % Uncomment for camera-ready version, but NOT for submission.
\begin{document}

\maketitle

\footnotetext[1]{Our code is available at \url{https://github.com/krafton-ai/lexico}.}

\begin{abstract}

We introduce \algoname, a novel KV cache compression method that leverages sparse coding with a universal dictionary. Our key finding is that key-value cache in modern LLMs can be accurately approximated using sparse linear combination from a small, input-agnostic dictionary of $\sim$4k atoms, enabling efficient compression across different input prompts, tasks and models.
Using orthogonal matching pursuit for sparse approximation, \algoname achieves flexible compression ratios through direct sparsity control. On GSM8K, across multiple model families (Mistral, Llama 3, Qwen2.5), \algoname maintains 90-95\% of the original performance while using only 15-25\% of the full KV-cache memory, outperforming both quantization and token eviction methods. Notably, \algoname remains effective in low memory regimes where 2-bit quantization fails, achieving up to 1.7× better compression on LongBench and GSM8K while maintaining high accuracy.

\end{abstract}

\begin{figure}[hb!]
    \centering
    \includegraphics[width=\linewidth]{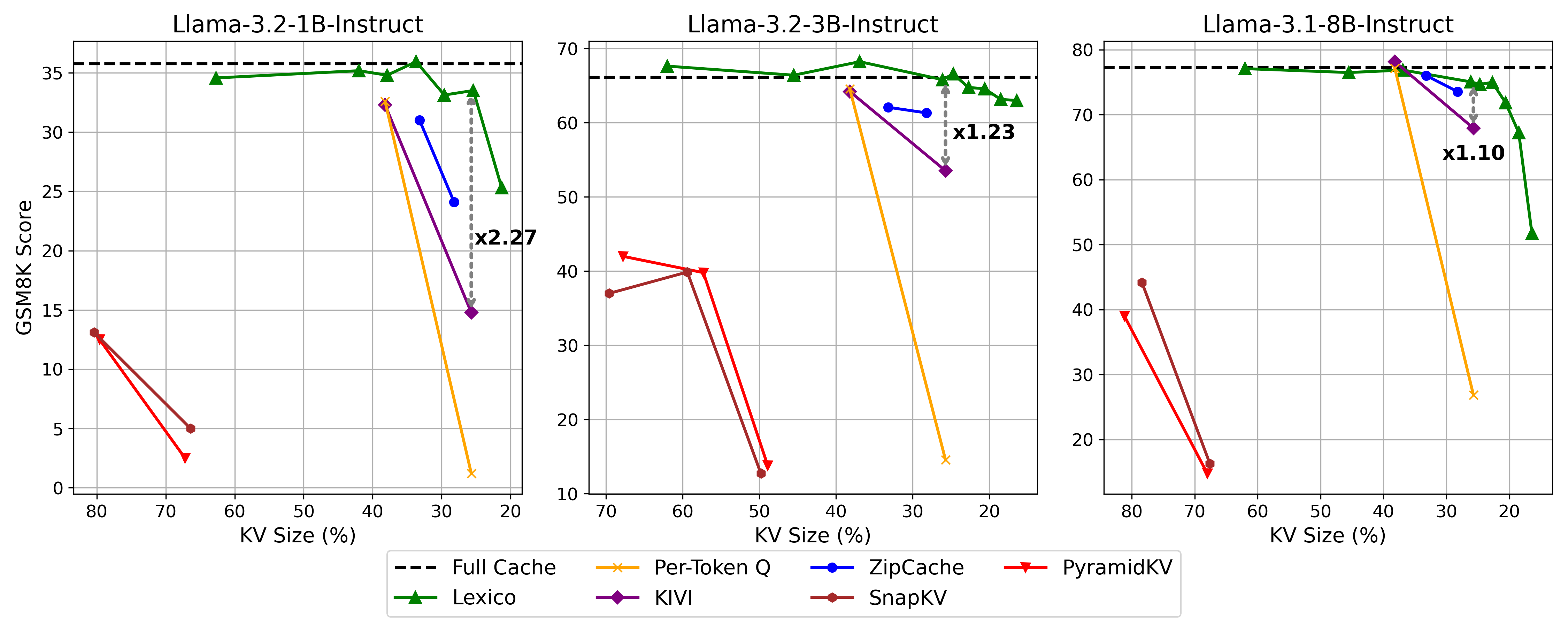}
    \caption{
        \new{\textbf{Memory usage vs. performance of \algoname compared to other \gls{kv} cache compression methods on GSM8K.}
        The figure illustrates the relationship between \gls{kv} cache size and the performance of \algoname on Llama models on GSM8K 5-shot evaluation. For \algoname, we use a dictionary size of $N = 4096$ atoms and keep the last $128$ tokens in full-precision (buffer size $n_b=128$). \algoname consistently outperforms both eviction-based methods (SnapKV, PyramidKV) and  quantization-based methods (per-token quantization, KIVI, ZipCache).} 
    }
    \label{fig:performance_vs_memory_usage_gsm8k}
\end{figure}

\section{Introduction}

Transformers~\citep{Vaswani2017Attention} have become the backbone of frontier \glspl{llm}, driving progress in domains beyond natural language processing. 
However, Transformers are typically limited by their significant memory requirements. 
This stems not only from the large number of model parameters,
but also from the having to maintain the \gls{kv} cache that grows proportional to the model size (i.e., the number of layers, heads, and also embedding dimension) and token length of the input.
Additionally, serving each model session typically  requires its own \gls{kv} cache, limiting opportunities for reuse across different user inputs, with the exception of prompt caching that only works for identitcal input prefixes.
This creates a bottleneck in generation speed for GPUs with limited memory~\citep{yu2022orca} and
thus, it has become crucial to alleviate \gls{kv} cache memory usage while preserving its original performance across domains.

\gls{kv} cache optimization research has explored both training-stage optimizations~\citep{shazeer2019fast, dai2024deepseekmoe, sun2024you} and post-training, deployment-focused methods~\citep{kwon2023efficient, lin2024infinite, ye2024chunkattention} to improve the efficiency of serving \glspl{llm}. 
Architectural approaches such as \gls{gqa}~\citep{ainslie2023gqa} aim to reduce the number of \gls{kv} heads, effectively reducing the size of the \gls{kv} cache.
% , a technique Llama-3~\citep{dubey2024llama} and Mistral models~\citep{jiang2023mistral} use.
However, most of these methods are not directly applicable as off-the-shelf methods to reduce \gls{kv} cache for pretrained \glspl{llm}, \new{leading to computationally costly, post-training compression efforts.}

Post-training approaches include selectively retaining certain tokens~\citep{Beltagy2020Longformer, xiao2023efficient, zhang2024h2o} 
\new{ and quantization methods, which have had empirical success when quantizing \gls{kv} cache into 2 or 4 bits~\citep{liu2024kivi, he2024zipcache, kang2024gear}.
However, eviction strategies have limitations on long-context tasks that require retaining a majority of previous tokens, while quantizations to 2 or 4 bits have clear upper bounds on compression rates.
}

In this paper, we focus on utilizing low-dimensional structures for efficient \gls{kv} cache compression. Prior work reports that each key vector lies in a low-rank subspace~\citep{singhania2024loki, wang2024model, yu2024effectively}. Yet, it is unclear if {\em all vectors} lie in the same subspace; if so, such redundancy remains to be taken advantage of. Thus, we naturally ask the following questions:
\vspace{-1mm}
\begin{center}
{\em 
Do keys and values lie in a low-dimensional subspace that is constant accross input sequences?\\
If so, can we leverage this for efficient KV cache compression?
}
\end{center}

\begin{figure}[t]
    \centering
    \begin{subfigure}{0.4\textwidth}
        \centering
        \includegraphics[width=\textwidth]{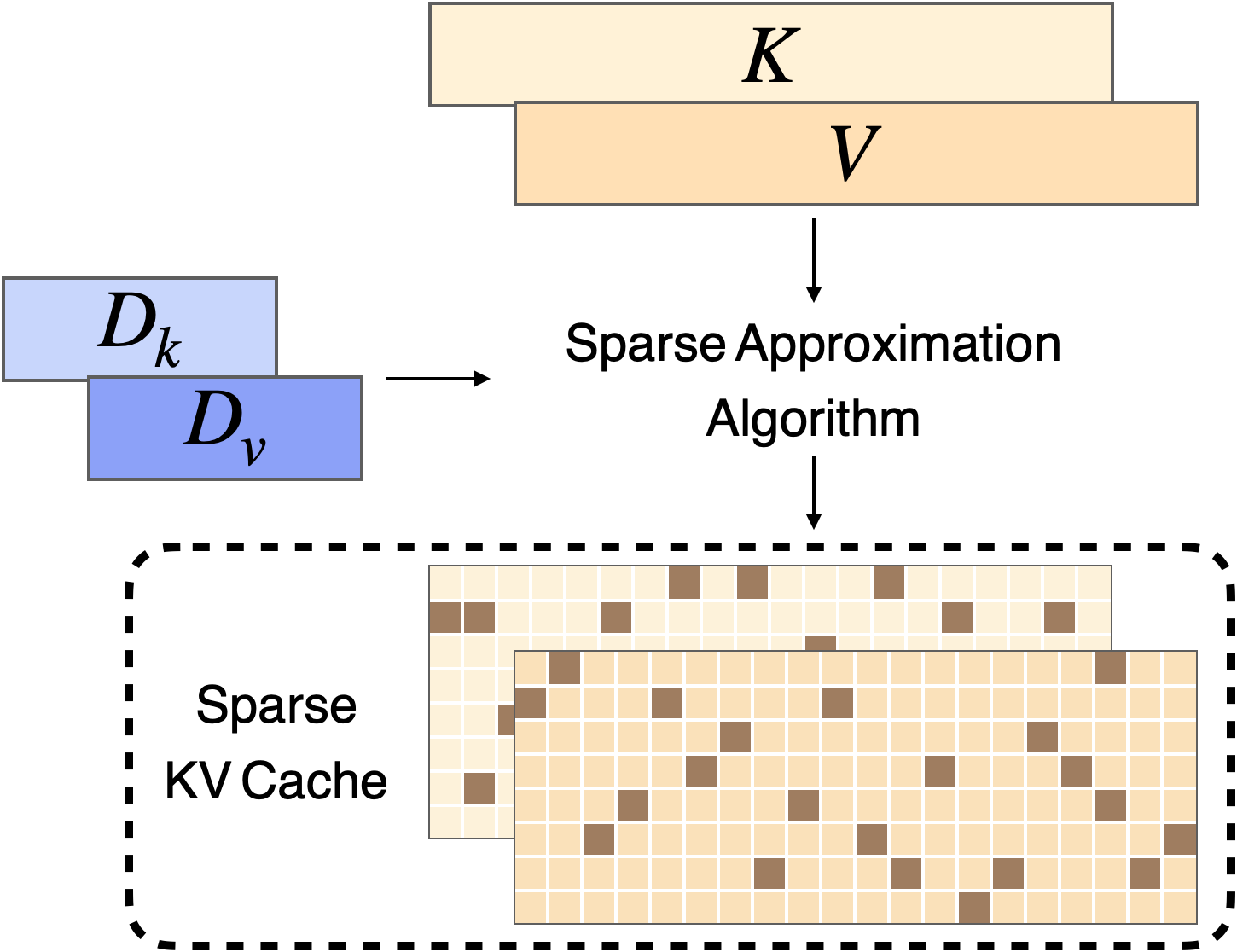}
        \caption{Prefilling Stage}
        \label{fig:prefilling_stage}
    \end{subfigure}
    \hfill
    \begin{subfigure}{0.58\textwidth}
        \centering
        \includegraphics[width=\textwidth]{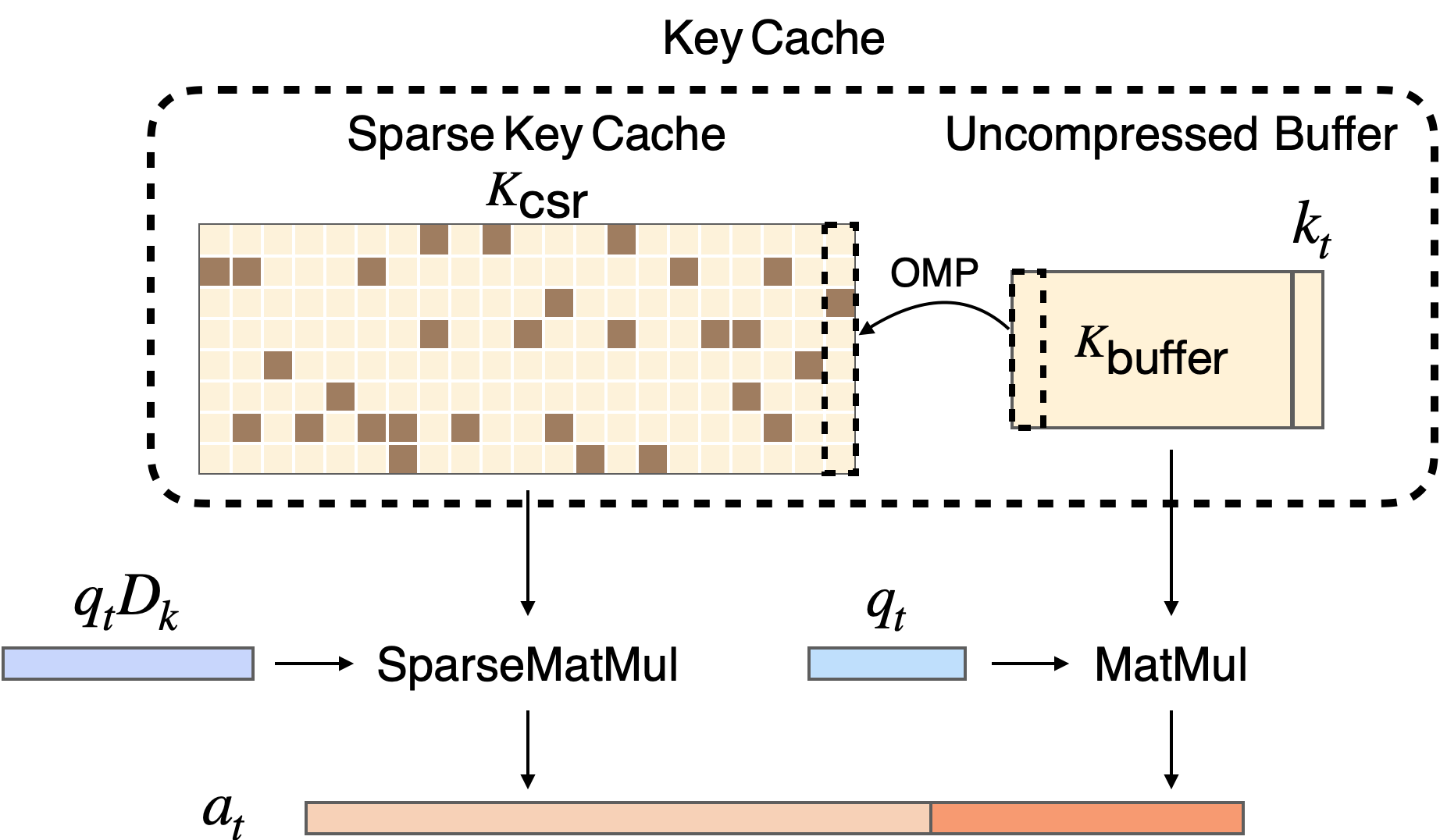}
        \caption{Decoding Stage}
        \label{fig:decoding_stage}
    \end{subfigure}
    \caption{\textbf{(a) Prefilling:} Following attention computation, \algoname uses OMP to find sparse representations of the \gls{kv} vectors ($3\text{-}8\times$ smaller). 
    \textbf{(b) Decoding:} Key cache consists of the compressed sparse key cache, $\mK_{\text{csr}}$, and an full-precision buffer, $\mK_{\text{buffer}}$, for the most recent tokens. $\vq_t$, $\vk_t$ represent the query, key vectors for the newly generated token. Computation is reduced by computing the query-dictionary product, $\vq_t \mD_k$, then multiplying $\mK_{\text{csr}}$, to get the pre-softmax attention score.}
    \label{fig:kv_dict_overview}
    \vspace{-2mm}
\end{figure}

Towards this end, we propose \algoname, a universal dictionary that serves as an overcomplete basis, which can sparsely decompose and reconstruct the \gls{kv} cache with sufficiently small reconstruction error that can be directly controlled via the level of sparsity of each reconstruction.

In Section~\ref{sec:sparse_approx}, we report our observation that a subset of key vectors cluster near each other, even though the keys are from different inputs, while some cluster on different subspaces.
To take advantage of such a low-dimensional structure, we draw inspiration from compressed sensing and  dictionary learning, areas of statistical learning and signal processing that  developed algorithms for information compression across various applicaiton domains~\citep{candes2006robust, donoho2006compressed, dong2014compressive, metzler2016denoising}.

\algoname is simple to learn, can be applied off-the-shelf for \gls{kv} cache compression, and only occupies small constant memory regardless of input or batch size. 
\new{
Methodologically, \algoname utilizes both sparsity-based compression (steps 1 and 2) and quantization (step 3) in three straightforward steps:
\begin{enumerate}
    \item \textbf{Dictionary pretraining:} 
    % As in Figure~\ref{fig:sparse_decoder},
    For our experiments, we train a dictionary on WikiText-103~\citep{merity2016wikitext} for each model. 
    This dictionary is only trained once and used universally across all tasks. It only occupies constant memory and does not increase with batch size. We note that this dictionary can be trained from richer sources to improve the overall performance of our sparse approximation algorithms.
    \item \textbf{Sparse decomposition:} 
    During prefilling and decoding (Figure~\ref{fig:kv_dict_overview}), \algoname decomposes key-value pairs into a sparse linear combination, which consists of $s$ pairs of reconstruction coefficients and dictionary indices pointing. This step by itself provides high compression rates.
    \item \textbf{Lightweight sparse coefficients:} 
    We obtain higher \gls{kv} cache compression rates by representing the sparse coefficients in 8 bits instead of \texttt{FP16}. Lowering precision to 8 bits yields minimal degradation.
    \algoname theoretically allows us to compress more than 2-bit quantization ($1/8$ of \texttt{FP16} \gls{kv} cache size) if $s\le 10$ when head dimension is $128$.
\end{enumerate}
}

Overall, we make the following contributions:
\begin{itemize}
    \item \textbf{Near-lossless performance:} Given similar memory requirements, \algoname performs on par with or better than baseline quantization methods on challenging language tasks, such as LongBench~\citep{bai2023longbench} and GSM8K~\citep{cobbe2021training}.
    \item \textbf{Compression rates beyond 2-bits:} \algoname's sparsity parameter enables both wider and more fine-grained control over desired memory usage. This allows us to explore performance when using under 15-20\% of the original \gls{kv} cache size, a low-memory regime previous compression methods could not explore.
    \item \textbf{Universality:} Instead of an input-dependent dictionary, we find a sufficiently small universal dictionary (per model) that can be used for all tasks and across multiple users. Advantageously, such dictionary does \textit{not} scale with batch size and can be used off-the-shelf.
\end{itemize}
\vspace{-1mm}

\section{Related Work}

Prior work on KV cache optimization has explored both training-stage and deployment-focused strategies to improve the efficiency of LLMs.
On the deployment side, \citet{kwon2023efficient} introduce a Paged Attention mechanism and the popular vLLM framework, which adapt CPU-style page memory to map KV caches onto GPU memory using a mapping table, thereby minimizing memory fragmentation and leveraging custom CUDA kernels for efficient inference. 
While there is a significant and important line of research in this direction~\citep{lin2024infinite, qin2024mooncake}, this direction is orthogonal to our work and can often be used in tandem with quantization.

Current post-training KV cache compression methods can broadly be categorized into eviction, quantization, and merging.
\citet{zhang2024h2o} introduced H2O, which uses attention scores to selectively retain tokens while preserving recent ones that are strongly correlated with current tokens.
Multiple works discuss various heuristics and algorithms to find which tokens can be discarded, while some works find how to complement evictions methods~\citep{ge2023model, li2024snapkv, liu2024scissorhands, devoto2024simple, dong2024less}.
For this line of work, there is a chance that evictions can work well together with \algoname, as \citet{liu2024scissorhands} have successfully combined quantization and eviction.

Quantization methods have also played a crucial role in reducing KV cache size without compromising model performance. Although there is a flurry of work, we only mention those that are most relevant to our discussion and methodology.
\citet{hooper2024kvquant} identified outlier channels in key matrices and developed KVQuant, while \citet{liu2024kivi} pursue a similar per-channel strategy in KIVI. 
Further extending these ideas, \citet{yue2024wkvquant} presented WKVQuant, which quantizes model weights as well as KV cache using two-dimensional quantization.
\citet{kang2024gear} follow similar per-channel key and per-token value quantization as KIVI, but with additional low-rank and sparse structures to manage quantization errors.

\section{KV Cache Compression with Dictionaries}

\subsection{Background \& Notation}

During autoregressive decoding in Transformer, the key and value states for preceding tokens are independent of subsequent tokens. 
As a result, these key and value states are cached to avoid recomputation, thereby accelerating the decoding process.

Let the input token embeddings be denoted as $\mX \in \sR^{l_{\text{seq}} \times d}$, where $l_{\text{seq}}$ and $d$ are the sequence length and model hidden dimension, respectively. For simplicity, we focus on a single layer and express the computation of query, key, and value states at each attention head during the prefilling stage as:
\begin{equation}
    \mQ^{(h)} = \mX \mW_q^{(h)}, \quad \mK^{(h)} = \mX \mW_k^{(h)}, \quad \mV^{(h)} = \mX \mW_v^{(h)},
\end{equation}
where $\mW_q^{(h)}, \mW_k^{(h)}, \mW_v^{(h)} \in \mathbb{R}^{d \times m}$ are the model weights with $m$ representing the head dimension.

Let $t$ represent the current step in the autoregressive decoding, and let $\vx_t \in \mathbb{R}^{1 \times d}$ denote the embedding of the newly generated token. The \gls{kv} cache up to but not including the current token, are denoted as $\mK_{t-1}^{(h)}$ and $\mV_{t-1}^{(h)}$, respectively. The typical output computation for each attention head $\vh_t^{(h)}$ using the \gls{kv} cache can be expressed as:
\begin{equation} \label{eq:attention_with_kv_cache}
    \vh_t^{(h)} = \mathrm{Softmax} \left( \vq_t^{(h)} \left( \mK_{t-1}^{(h)} \parallel \vk_t^{(h)} \right)^\top /  {\sqrt{m}} \right) \left( \mV_{t-1}^{(h)} \parallel \vv_t^{(h)} \right), 
\end{equation}
where $\vq_t^{(h)}, \vk_t^{(h)}, \vv_t^{(h)}$ represent the query, key, and value vectors for the new token embedding $\vx_t$. Here, $\parallel$ denotes concatenation along the sequence length dimension.

\subsection{Sparse Approximation}
\label{sec:sparse_approx}

Given a dictionary, our goal is to decompose and represent \gls{kv} cache efficiently,
\textit{i.e.}, approximate a vector $\vk \in \mathbb{R}^m$ as a linear combination of a few vectors (atoms) from an overcomplete dictionary $\mD \in \mathbb{R}^{m \times N}$. 
This reconstruction is given by $\vk = \mD \vy$, where $\vy \in \mathbb{R}^N$ is the sparse representation vector such that $s = || \vy ||_0$.
For implementation, $\vy$ only requires space proportional to $s$, not $N$.

\begin{figure}[h]
    \centering
    \begin{subfigure}{0.4\textwidth}
        \centering
        \includegraphics[width=\textwidth]{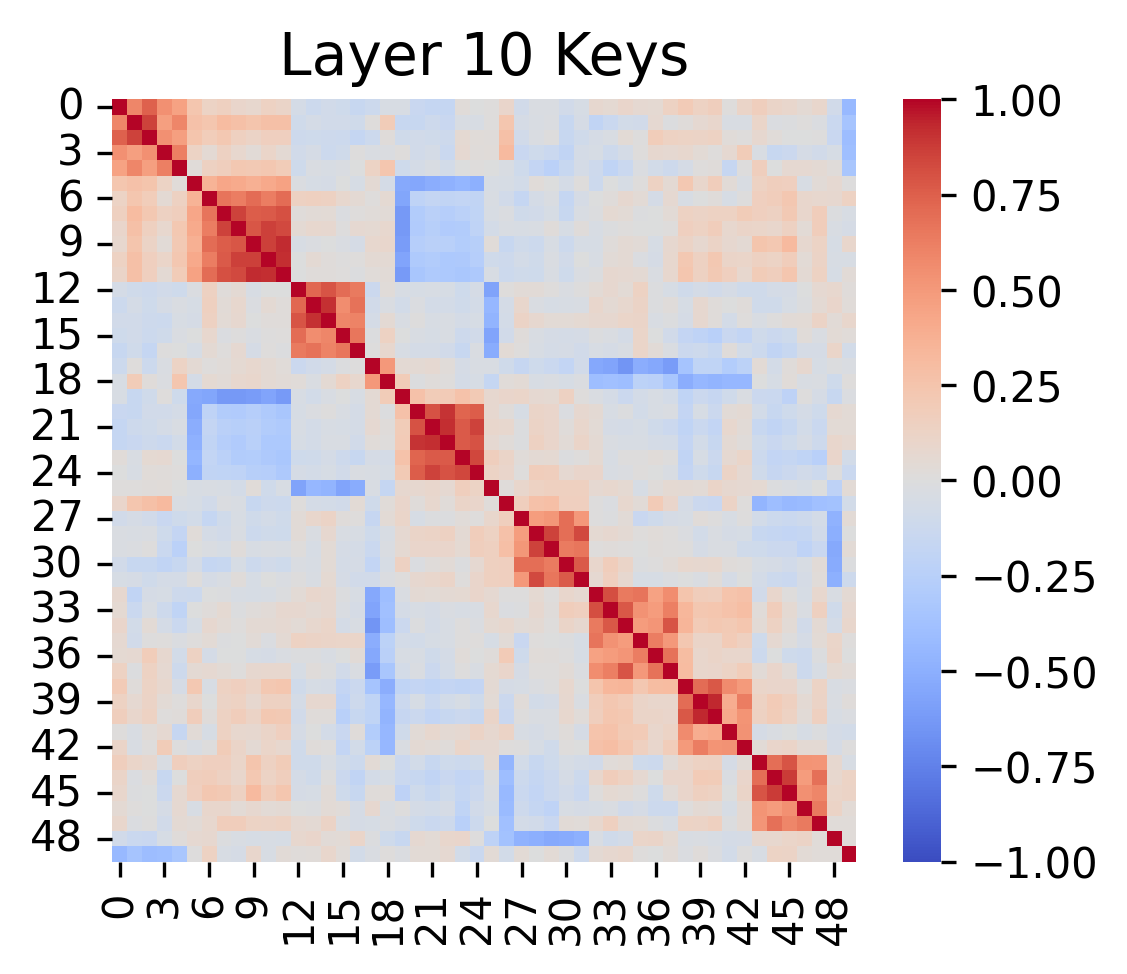}
    \end{subfigure}
    \hspace{0.02\textwidth} % Adjust this value as needed to control the spacing
    \begin{subfigure}{0.4\textwidth}
        \centering
        \includegraphics[width=\textwidth]{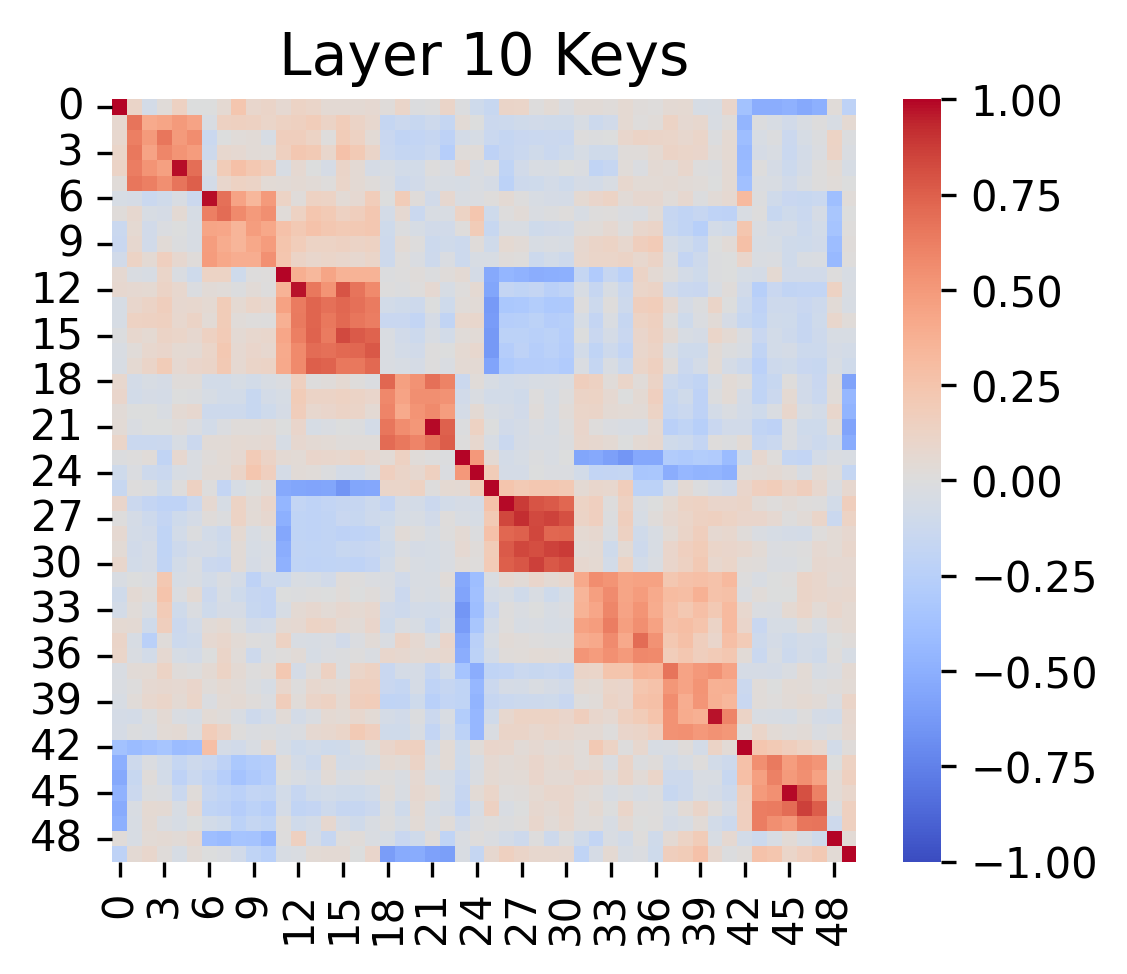}
    \end{subfigure}
    \caption{\textbf{Left} shows a pairwise cosine similarity matrix between key vectors generated from one input text from all heads in Layer 10 of Llama-3.1-8B-Instruct. Keys are sorted by similarity to demonstrate the clusters. \textbf{Right} shows the similarity matrix between key vectors from two \textit{different} input texts. These plots indicate that there may exist a mixture of low-dimensional subspaces in the space of \textit{all} possible keys, a hypothesis that naturally leads to dictionary learning.}
    \label{fig:clustering_of_keys}
\end{figure}

We hypothesize that the \gls{kv} cache, like other domains where sparse approximation is effective, contains inherent redundancy that can be leveraged for efficient compression. 
For instance, Figure~\ref{fig:clustering_of_keys} presents pairwise cosine similarity plots for keys generated during inference on a random subset of the WikiText dataset. 
Here, we observe that key vectors cluster in multiple different subspaces.
Dictionary learning can take advantage of such redundancy, enabling \gls{kv} vectors to be represented by a compact set of atoms with only a few active coefficients. 

Sparse approximation, which aims to find $\vy$ with minimum sparsity given $\vk$ and $\mD$, while ensuring a small reconstruction error, is NP-hard. This optimization problem is typically formulated as:
\begin{equation}
    \underset{\vy}{\text{min}} \left\| \vy \right\|_0 \text{subject to}  \left\| \vk - \mD \vy \right\|_2 \leq \delta \left\| \vk \right\|_2  \text{ for some relative error threshold } \delta > 0
\end{equation}

In this work, we adopt \gls{omp} as the sparse approximation algorithm. Given an input key or value vector $\vk$, a dictionary $\mD$, and a target sparsity $s$, \gls{omp} iteratively selects dictionary atoms to minimize the $\ell_2$-reconstruction error, with the process continuing until the specified sparsity $s$ is reached.
% Algorithm~\ref{alg:omp_simple} illustrates a basic version of \gls{omp} for clarity. 
% \jongho{Add error thresholding}
Our implementation of \gls{omp} builds on advanced methods that utilize properties of the Cholesky inverse~\citep{zhu2020efficient} to optimize performance. Additionally, we incorporate implementation details from \cite{lubonja2024efficient} for efficient batched GPU execution and extend it to include an extra batch dimension, allowing for parallel processing across multiple dictionaries. The full algorithm is detailed in Appendix~\ref{sec:omp_implementation_details}.

\subsection{Learning Layer-specific Dictionaries} \label{sec:learning_layer_specific_kv_dict}

\paragraph{Layer-specific dictionaries.}
While the sparse approximation algorithm is crucial, achieving a high compression ratio relies heavily on well-constructed dictionaries. In this section, we describe the process for training the dictionaries used in \algoname.
We adopt distinct dictionaries for the key and value vectors in each transformer layer due to their different functionalities. 
We denote the key and value dictionaries at each layer as $\mD_k$ and $\mD_v \in \mathbb{R}^{m \times N}$, where $N$ is the fixed dictionary size.
With $N = 1024$, the dictionaries add an additional 16.8MB to the model's storage requirements for 7B/8B models.

\begin{figure}[t]
    \centering
    \includegraphics[width=0.7\textwidth]{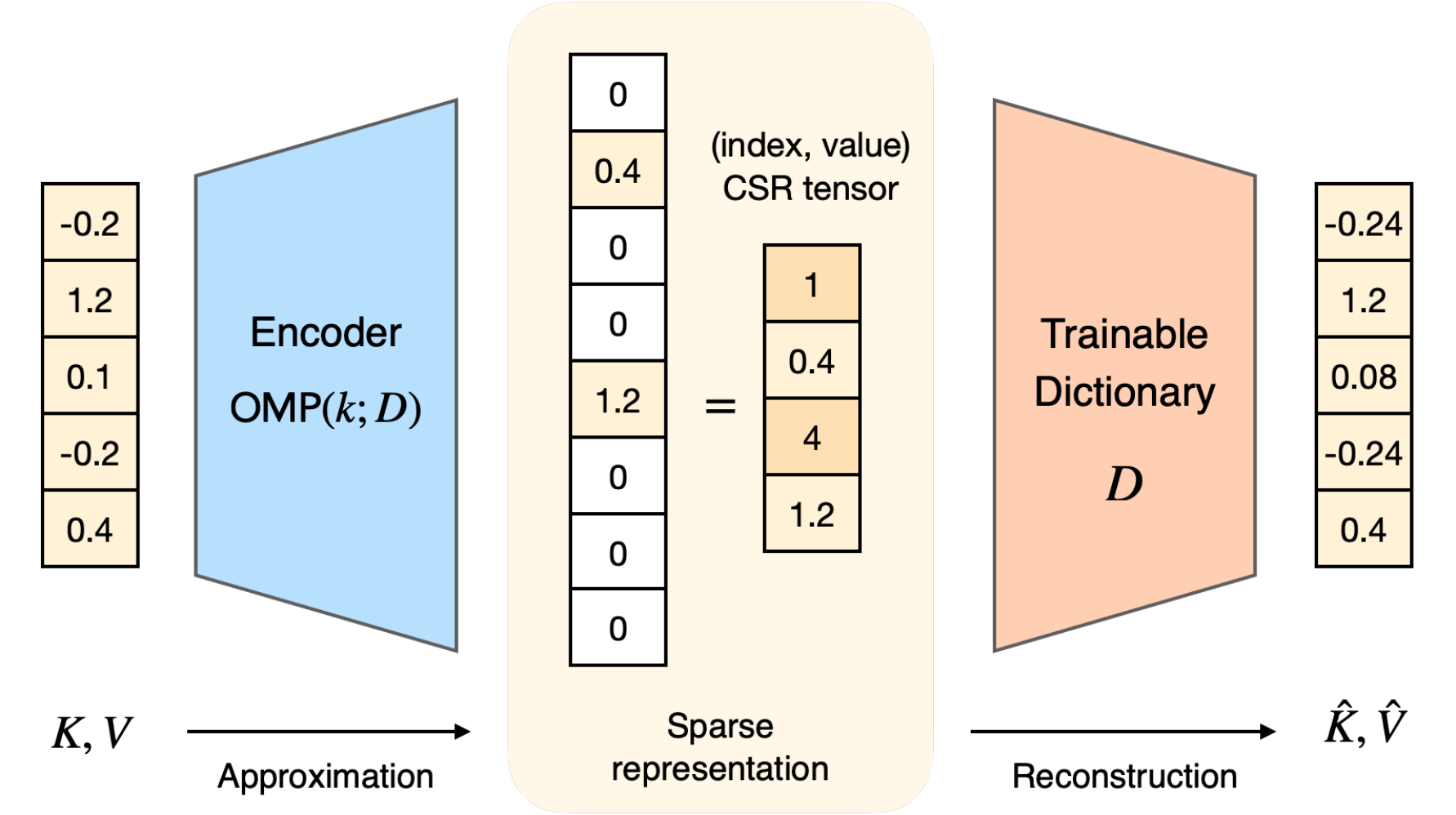}
    \caption{\textbf{Dictionary Learning of \algoname.} We train a linear layer $\mD$ (our dictionary) that minimizes $\ell_2$-reconstruction error of \gls{kv} cache. The \gls{kv} cache of layer $i$ are used as training data for dictionary $\mD^{(i)}$. Each step, we apply \gls{omp} with fixed $\mD$ to represent \gls{kv} as a vector of sparse coefficients; we then perform a step of gradient descent on $\mD$ and repeat the process.
    A sparse vector can be efficiently stored as a \gls{csr}, using a tuple of 16-bit index and 8-bit value.}
    \label{fig:sparse_decoder}
    % \vspace{-2mm}
\end{figure}

As shown in Figure~\ref{fig:sparse_decoder}, we train layer-specific \gls{kv} dictionaries through direct gradient-based optimization. For a given key or value vector, denoted as $\vk \in \mathbb{R}^{m}$ and a dictionary $\mD \in \mathbb{R}^{m \times N}$, the \gls{omp} algorithm approximates the sparse representation $\vy \in \mathbb{R}^{N}$. This process is parallelized across multiple dictionaries, but for simplicity, we present the notation for a single dictionary. The dictionary training objective minimizes the $\ell_2$ norm of the reconstruction error, with the loss function $\mathcal{L} = \left\| \vk - \mD \vy \right\|_2^2$.
We enforce unit norm constraints on the dictionary atoms by removing any gradient components parallel to the atoms before applying updates.

\paragraph{Training.}
The dictionaries are trained on \gls{kv} pairs generated from the WikiText-103 dataset using Adam~\citep{kingma2014adam} with a learning rate of 0.0001 and a cosine decay schedule over 20 epochs. 
The dictionaries are initialized with a uniform distribution, following the default initialization method for linear layers in PyTorch.
For Llama-3.1-8B-Instruct, with a sparsity of $s=32$ and a dictionary size of $N=1024$, the training process takes about 2 hours on an NVIDIA A100 GPU.

\begin{table}[hbp]
    \centering
    \footnotesize
    \caption{\textbf{Reconstruction error.} Relative reconstruction errors of different methods when training dictionary of size $1024$ and sparsity $s=32$ on WikiText-103. Sparse Autoencoder is a two-layer perceptron with hard top-$k$ thresholding as activation (encoder as a linear layer + activation in Figure~\ref{fig:sparse_decoder}). 
    \algoname is optimized using \gls{omp} as encoder. \gls{kv} cache is generated from Llama-3.1-8B-Instruct.}
    \begin{tabular}{lccc}
        \toprule
        \textbf{Test Dataset}      & \textbf{\algoname}      & \textbf{Sparse Autoencoder}     & \textbf{Random Dictionaries} \\
        \midrule
        WikiText-103          & $\mathbf{0.17} \pm 0.06$     & $0.20 \pm 0.05$     & $0.27 \pm 0.02$  \\
        CNN/DailyMail         & $\mathbf{0.19} \pm 0.05$     & $0.22 \pm 0.04$     & $0.27 \pm 0.02$  \\
        IMDB                  & $\mathbf{0.18} \pm 0.05$     & $0.22 \pm 0.05$     & $0.27 \pm 0.02$  \\
        TweetEval             & $\mathbf{0.18} \pm 0.06$     & $0.21 \pm 0.05$     & $0.27 \pm 0.02$  \\
        \bottomrule
    \end{tabular}
    \label{tab:dictionary_comparison}
\end{table}

We demonstrate our trained dictionaries reconstruct and generalize better than dictionaries trained using sparse autoencoders (similarly to those from \citet{makhzani2013k, bricken2023towards}) across several corpora in Table~\ref{tab:dictionary_comparison}.
Our method consistently achieves lower relative reconstruction errors, such as $0.19 \pm 0.05$ on out-of-domain dataset CNN/DailyMail, and this trend is consistent across other datasets. 

Despite being trained only on WikiText-103, \algoname dictionaries demonstrate a degree of universality:
our dictionaries achieve lower test loss on out-of-domain datasets such as TweetEval than the test loss on WikiText-103 for sparse autoencoders, offering significant compression with minimal reconstruction error.
In the next subsection, we explore how low $\ell_2$-reconstruction loss translates to strong performance preservation in language modeling.

\subsection{Prefilling and Decoding with \algoname}

During the prefilling stage, each layer generates the \gls{kv} vectors for the input tokens, as illustrated in Figure~\ref{fig:prefilling_stage}. \algoname uses full-precision \gls{kv} vectors for attention computation, which are then passed to subsequent layers. Subsequently, \gls{omp} finds the sparse representations of the \gls{kv} vectors using layer-specific key and value dictionaries, $\mD_k$ and $\mD_v$.

% The sparse representations are stored in the \gls{csr} format, where each row corresponds to a single key or value vector. In \gls{csr}, the nonzero values are stored along with their corresponding dictionary indices, and an offset array indicates the starting position of each vector, allowing for variable sparsity levels among different rows. We use FP8 with E4M3 encoding for values and int16 for indices. For a given sparsity level $s$, the size of the sparse representation is given by $s \times 1 + s \times 2 + 2 = 3s + 2$ bytes.
% Considering a head dimension of $128$, a single uncompressed key or value vector, using FP16 precision, takes up $128 \times 2 = 256$ bytes, giving the memory percentage of $\frac{4s + 2}{256} \times 100 \%$.

The compressed key and value caches are denoted as $\mK_{\text{csr}}, \mV_{\text{csr}} \in \mathbb{R}^{l_{\text{seq}} \times N}$ and replace the full-precision \gls{kv} cache. The reconstruction of the \gls{kv} cache is performed as follows:
\begin{equation}
    \hat{\mK} = \mK_{\text{csr}} \mD_k^\top, \quad \hat{\mV} = \mV_{\text{csr}} \mD_v^\top
\end{equation}

Recall that at the $t$-th iteration of autoregressive decoding, each layer receives $\vq_t$, $\vk_t$, and $\vv_t$, the query, key, and value vectors corresponding to the newly generated token. Similarly to prior work~\citep{liu2024kivi, kang2024gear}, we find that keeping a small number of recent tokens in full precision improves the generative performance of the model. To achieve this, we introduce a buffer that temporarily stores recent tokens in an uncompressed state. The \gls{kv} vectors stored in the buffer are denoted as $\mK_{\text{buffer}}, \mV_{\text{buffer}} \in \mathbb{R}^{n_b \times m}$, where $n_b$ is the number of \gls{kv} vectors in the buffer. 
The key cache up to, but not including the new token at iteration $t$, is then reconstructed as follows:
\begin{equation}
    \hat{\mK}_{t-1} = \mK_{\text{csr}} \mD_k^\top \parallel \mK_{\text{buffer}}
\end{equation}

Substituting this reconstruction into the Equation~\ref{eq:attention_with_kv_cache}, the attention weights for each head $\va_t^{(h)}$ are computed as:
\begin{equation}
    \va_t^{(h)} = \text{Softmax} \left( \vq_t^{(h)} \Large( \mK_{\text{csr}}^{(h)} \mD_k^\top \parallel \mK_{\text{buffer}}^{(h)} \parallel \vk_t^{(h)} \Large)^\top / \sqrt{m} \right)
\end{equation}

A key implementation 
% detail of \algoname is that it avoids the need to reconstruct the key cache for the entire sequence. Instead, 
is that attention for the compressed sparse key cache and the uncompressed key cache is computed separately. 
For compressed sparse key cache, we first compute the product $\vq_t^{(h)} \mD_k$ before we multiply
$\mK_{\text{csr}}$, directly calculating the pre-softmax attention scores for compressed tokens.
Attention for the buffer tokens is computed as usual. These scores are then concatenated with softmax to produce the final attention weights (Figure~\ref{fig:decoding_stage}). This process is formalized as:
\begin{equation}
    \va_t^{(h)} = \text{Softmax} \left( \left(  \vq_t^{(h)} \mD_k \mK_{\text{csr}}^{(h)\top} \mid \vq_t^{(h)} \large( \mK_{\text{buffer}}^{(h)} \parallel \vk_t^{(h)} \large)^\top \right) / {\sqrt{m}} \right),
\end{equation}
where $\mid$ represents concatenation along columns for attention scores.

When the buffer reaches capacity, 
\gls{omp} compresses the \gls{kv} vectors for the oldest $n_a$ tokens in the buffer. This process is independent of the attention computation for the newest token and can therefore be performed in parallel.

\paragraph{Time and space complexity.}

The sparse representations are stored in \gls{csr} format, with values encoded in \texttt{FP8(E4M3)}, and all indices, including offsets, are stored as \texttt{int16}. Each row in \gls{csr} corresponds to a single key or value vector. For a given sparsity level $s$, the memory usage includes: nonzero values ($s$ bytes), dictionary indices ($2s$ bytes), and the offset array (2 bytes), resulting in a total size of $3s + 2$ bytes. For a head dimension of 128, a fully uncompressed vector using \texttt{FP16} takes $256$ bytes, yielding a memory usage of $\frac{3s + 2}{256} \times 100 \approx 1.17s \%$ (\emph{e.g.}, $37.5\%$ for $s=32$).

In terms of time complexity, computing $\vq_t \mK_t^\top$ for a single head requires $O(l_{\text{seq}} m)$ multiplications. 
On the other hand, $\vq_t \mD_k \mK_{\text{csr}}^\top$ needs $O(Nm + l_{\text{seq}} s)$ multiplications. This means that our computation is particularly well-suited for long-context tasks when $l_{\text{seq}} > m$ where $m$ is anywhere between 1024 and 4096. For short contexts when $l_{\text{seq}} < m$, our method only adds a small overhead to attention computation in actuality.
\section{Experiments}

% \subsection{Settings}

\paragraph{Setup.}
We evaluate our method on various models (Llama-3-8B, Llama-3.1-8B-Instruct, Llama-3.2-1B-Instruct, Llama-3.2-3B-Instruct, Mistral-7B-Instruct),
using dictionaries trained on WikiText-103, as done in Section~\ref{sec:learning_layer_specific_kv_dict}.
To assess the effectiveness of \algoname in memory reduction while maintaining long-context understanding, we conduct experiments on selected tasks from LongBench~\citep{bai2023longbench}, following the setup of \citet{liu2024kivi}. 
See Table~\ref{tab:longbench_task_statistics} in Appendix~\ref{sec:longbench_task_statistics} for task details.

Additionally, we evaluate generative performance on complex reasoning tasks, such as GSM8K~\citep{cobbe2021training} with 5-shot prompting
and MMLU-Pro Engineering/Law~\citep{wang2024mmlu} with zero-shot chain-of-thought.
We choose these MMLU-Pro subjects since they are deemed the most difficult as they require complex formula derivations or deep understanding of legal knowledge intricacies.
We compare our method against
two kinds of \gls{kv} cache compression methods: namely, quantization-based compression and eviction-based compression.
For quantization-based methods, we evaluate KIVI~\citep{liu2024kivi}, ZipCache~\citep{he2024zipcache}, and the Hugging Face implementation for per-token quantization.
For eviction-based methods, we evaluate PyramidKV~\citep{cai2024pyramidkv} and SnapKV~\citep{li2024snapkv}.

% KIVI, a quantization method that quantizes keys per-channel and values per-token in groups of size $g$.
We refer to the 4-bit and 2-bit versions of KIVI as KIVI-4 and KIVI-2, respectively, and denote its quantization group size as $g$.

We report \gls{kv} size as the average percentage of the compressed cache relative to the full cache at the end of generation. 
\algoname's sparsity $s$ is set to match the \gls{kv} size of the baseline.

\paragraph{Hyperparameter settings.}
For both experiments, \algoname uses a dictionary size of $N=4096$, a buffer size of $n_b=128$, and an approximation window size $n_a=1$, compressing the oldest token with each new token generated.
For KIVI-4 and KIVI-2, we use a quantization group size of $g = 32$ and a buffer size of $n_b = 128$
, as is tested and recommended in \citet{liu2024kivi}, for LongBench.
For GSM8K and MMLU-Pro, we test for stronger memory savings, so we use $g = 64$ and $n_b = 64$ for KIVI.

\subsection{Experimental Results}

\paragraph{LongBench results.}
Table~\ref{tab:longbench_results} presents the performance of \algoname and KIVI on LongBench tasks. 
% using the Llama-3.1-8B-Instruct and Mistral-7B-Instruct models. 
\algoname 
demonstrates better performance than KIVI with similar or even smaller \gls{kv} sizes. 
Notably, \algoname enables exploration of extremely low memory regimes that KIVI-2 cannot achieve. 
At a memory usage of just $12.4\%$ \gls{kv} size, \algoname maintains reasonable long-context understanding, with only 5.6$\%p$ and 4.4$\%p$ performance loss on Llama-3.1-8B-Instruct and on Mistral-7B-Instruct-v0.3, respectively, compared to the full cache (\texttt{FP16}).
The largest performance loss comes from tasks with the lowest full cache accuracy, Qasper, yet there is almost no loss in simpler tasks, such as TriviaQA.
This indicates that difficult tasks that require more complex understanding are much more sensitive to performance loss.
Hence, it is important to evaluate on GSM8K, one of the harder natural language reasoning tasks, as we do next.

\begin{table}[tbp]
    \centering
    \scriptsize  % You can try \tiny if needed, but it may affect readability
    \setlength{\tabcolsep}{4pt}  % Adjust column padding (default is 6pt)
    \caption{\textbf{Experimental results on LongBench.} For \algoname, we use $N = 4096$ as the dictionary size and $n_b = 128$ as the buffer size. For KIVI, we use $g = 32$ (group size for quantization) and $n_b = 128$ (buffer size).
    Sparsity level $s$ is set to match average \gls{kv} size of KIVI, while $s=8$ corresponds to cache size unattainable by common 2-bit quantizations. Full cache is in \texttt{FP16}.}
    \label{tab:longbench_results}
    \begin{tabular}{@{}lccccccccccc@{}}
        \toprule
        \textbf{Method} & \textbf{KV Size} & \textbf{Qasper} & \textbf{QMSum} & \textbf{MultiNews} & \textbf{TREC} & \textbf{TriviaQA} & \textbf{SAMSum} & \textbf{LCC} & \textbf{RepoBench-P} & \textbf{Average} \\ 
        \midrule
        
        % Llama 3.1 8B Instruct Section
        \multicolumn{11}{c}{\textbf{Llama-3.1-8B-Instruct}} \\  % Model name spanning 11 columns
        Full Cache    & $100\%$ & $22.54$ & $24.57$ & $27.44$ & $72.5$ & $91.65$ & $43.47$ & $63.15$ & $56.76$ & $50.26$ \\
        \midrule
        KIVI-4              & $33.2\%$ & $\mathbf{22.83}$ & $23.72$          & $\mathbf{27.95}$ & $71.0$          & $90.39$          & $\mathbf{44.25}$ & $\mathbf{62.93}$ & $55.48$         & $49.78$ \\
        \algoname $_{s=24}$ & $30.6\%$ & $21.68$          & $\mathbf{24.25}$ & $27.20$          & $\mathbf{72.5}$ & $\mathbf{91.58}$ & $42.93$          & $62.92$          & $\mathbf{56.51}$ & $\mathbf{49.95}$ \\
        \midrule
        KIVI-2               & $21.1\%$ & $13.77$          & $22.72$          & $\mathbf{27.35}$ & $71.0$          & $90.85$          & $\mathbf{43.53}$ & $62.03$          & $53.00$     & $48.03$ \\
        \algoname $_{s=16}$  & $21.4\%$ & $\mathbf{15.45}$ & $\mathbf{23.13}$ & $25.78$          & $\mathbf{72.5}$ & $\mathbf{92.25}$ & $42.02$          & $\mathbf{63.01}$ & $\mathbf{55.58}$ & $\mathbf{48.71}$ \\
        \midrule
        \algoname $_{s=8}$  & $\mathbf{12.4\%}$ & $11.66$ & $21.04$ & $22.35$ & $60.0$ & $91.01$ & $40.30$ & $59.60$ & $51.46$ & $44.68$ \\
        
        \midrule
        
        % Mistral 7B Instruct Section
        \multicolumn{11}{c}{\textbf{Mistral-7B-Instruct-v0.3}} \\  % Model name spanning 11 columns
        Full Cache    & $100\%$ & $41.58$ & $25.69$ & $27.76$ & $76.0$ & $88.59$ & $47.58$ & $59.37$ & $60.60$ & $53.40$ \\
        \midrule
        KIVI-4              & $33.2\%$ & $40.37$          & $24.51$          & $\mathbf{27.75}$ & $74.0$          & $88.36$          & $\mathbf{47.56}$ & $58.49$          & $58.31$ & $52.42$ \\
        \algoname $_{s=24}$ & $30.6\%$ & $\mathbf{41.01}$ & $\mathbf{25.32}$ & $27.51$          & $\mathbf{76.0}$ & $\mathbf{88.84}$ & $46.27$          & $\mathbf{59.98}$ & $\mathbf{59.44}$ & $\mathbf{53.05}$ \\
        \midrule
        KIVI-2               & $21.1\%$ & $38.24$          & $24.08$          & $\mathbf{26.99}$ & $74.5$          & $88.34$          & $\mathbf{47.66}$ & $57.51$          & $56.46$ & $51.72$ \\
        \algoname $_{s=16}$  & $21.4\%$ & $\mathbf{40.34}$ & $\mathbf{24.97}$ & $26.36$          & $\mathbf{76.0}$ & $\mathbf{89.31}$ & $45.84$          & $\mathbf{59.31}$ & $\mathbf{59.50}$ & $\mathbf{52.70}$ \\
        \midrule
        \algoname $_{s=8}$  & $\mathbf{12.4\%}$ & $33.03$ & $22.80$ & $22.85$ & $68.5$ & $87.85$ & $43.10$ & $56.66$ & $56.85$ & $48.96$ \\
        
        \bottomrule
    \end{tabular}
\end{table}

% \vspace{5mm}
\paragraph{GSM8K results.}
The performance of \algoname on GSM8K compared to KIVI is shown in Table~\ref{tab:gsm_results}. 
With a \gls{kv} size of $36.9\%$, \algoname on Llama 8B models experiences a slight accuracy drop of less than $3\%p$, underperforming KIVI-4 at a similar \gls{kv} size.
However, in the lower memory regime near $25\%$ \gls{kv} size, \algoname significantly outperforms KIVI-2, achieving a higher accuracy by $8.2\%p$ on the Llama-3-8B model and $7.1\%p$ on the Llama-3.1-8B-Instruct model.
These results highlight the robustness of \algoname in low-memory settings, demonstrating that low reconstruction error can be achieved using only a few atoms from our universal dictionary. To further test the resilience of \algoname , we set the sparsity to $s = 4$, observing a noticeable drop in accuracy on the Llama-3.1-8B-Instruct model. Despite this, both Llama models maintain an accuracy above $40\%$, which is remarkable given that only 4 atoms from \algoname were used for each key-value vector, utilizing just $15.8\%$ of the full cache, including the buffer.

The performance of \algoname on the Mistral-7B-Instruct model is even more impressive.
We demonstrate that for Mistral, \algoname not only outperforms KIVI-4 and KIVI-2 but also achieves higher accuracy with even less memory usage.
We also evaluate \algoname with $s=4$ on the Mistral model and observe an accuracy of $39.2\%$, further demonstrating robustness in low-memory settings.

\begin{table}[tbp]
    \centering
    \small  % Adjust size as needed
    \caption{\textbf{Experimental results on GSM8K.} For \algoname, we use $N = 4096$ as the dictionary size and $n_b = 128$ as the buffer size. For KIVI, we use $g = 64$ (group size for quantization) and $n_b = 64$ (buffer size). 
    Sparsity level $s$ is set to match the average \gls{kv} size of KIVI, while $s=4$ corresponds to cache size unattainable by common 2-bit quantizations. Full cache is in \texttt{FP16}. We include example generations of KIVI and \algoname in Appendix~\ref{app:gsm8k_example}.}
    \label{tab:gsm_results}
    
    \begin{subtable}{0.48\linewidth}  % First subtable with its own width
        \centering
        \caption{Llama 3.x 8B Models}
        \begin{tabular}{@{}lccc@{}}
            \toprule
            \textbf{Method} & \textbf{KV Size} & \textbf{Llama-3-8B} & \textbf{3.1-8B-Instruct} \\ 
            \midrule
            Full Cache              & $100\%$       & $49.89$       & $79.61$                   \\
            \midrule
            KIVI-4                  & $38.2\%$      & $\mathbf{49.13}$       & $\mathbf{78.17}$                     \\
            \algoname $_{s=24}$     & $36.9\%$      & $48.29$       & $76.88$                       \\
            \midrule
            KIVI-2                  & $25.7\%$      & $40.56$       & $67.93$                     \\
            \algoname $_{s=14}$     & $26.1\%$      & $\mathbf{48.75}$& $\mathbf{75.06}$     \\
            \midrule
            \algoname $_{s=4}$      & $\mathbf{15.8\%}$     & $40.03$       & $51.71$        \\
            \bottomrule
        \end{tabular}
    \end{subtable}
    \hspace{0.07\linewidth}
    \begin{subtable}{0.43\linewidth}  % Second subtable with its own width
        \centering
        \caption{Mistral 7B v0.3 Model}
        \begin{tabular}{@{}lcc@{}}
            \toprule
            \textbf{Method} & \textbf{KV Size} & \textbf{7B-Instruct} \\ 
            \midrule
            Full Cache              & $100\%$       & $48.60$                   \\
            \midrule
            KIVI-4                  & $38.2\%$       & $48.52$                   \\
            \algoname $_{s=20}$     & $32.7\%$      & $\mathbf{48.60}$          \\
            \midrule
            KIVI-2                  & $25.7\%$      & $42.91$                   \\
            \algoname $_{s=10}$     & $22.0\%$      & $\mathbf{44.35}$       \\
            \midrule
            \algoname $_{s=4}$      & $\mathbf{15.8\%}$     & $39.20$       \\
            \bottomrule
        \end{tabular}
    \end{subtable}
\end{table}

\begin{figure}[htbp]
    \centering
    \begin{minipage}[t]{0.35\linewidth}
        \centering
        \includegraphics[width=\linewidth]{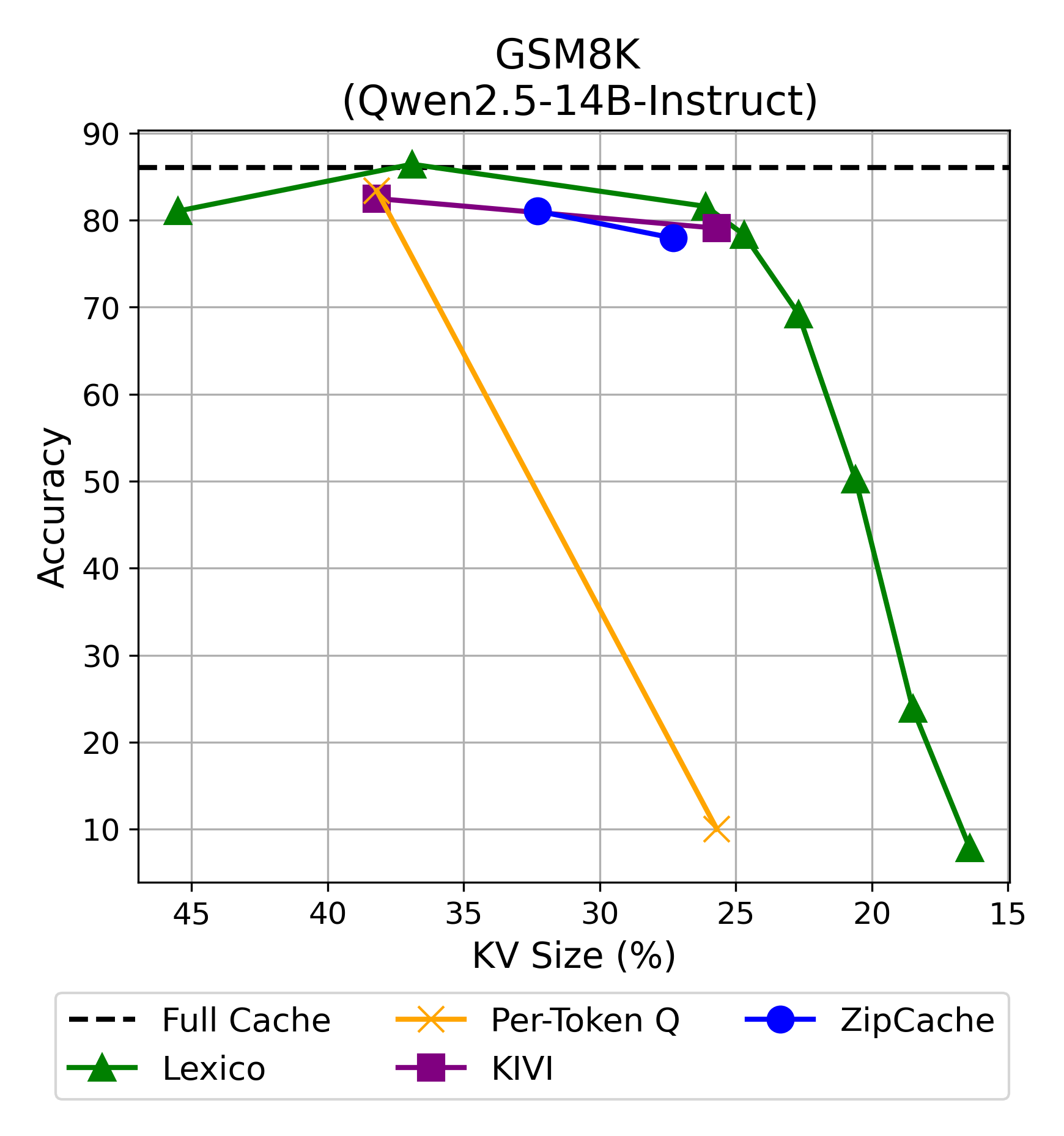}
        \caption{
            \textbf{Memory usage vs. performance of Qwen2.5-14B-Instruct with \algoname on GSM8K.} 
            We compare the performance of \algoname against quantization methods on Qwen2.5-14B-Instruct, with its weights quantized to 4 bits.
            For \algoname, we use $N = 4096$ as the dictionary size and $n_b = 128$ as the buffer size.
        }
        \label{fig:performance_vs_memory_usage_qwen_gsm8k}
    \end{minipage}
    \hfill
    \begin{minipage}[t]{0.62\linewidth}
        \centering
        \includegraphics[width=\linewidth]{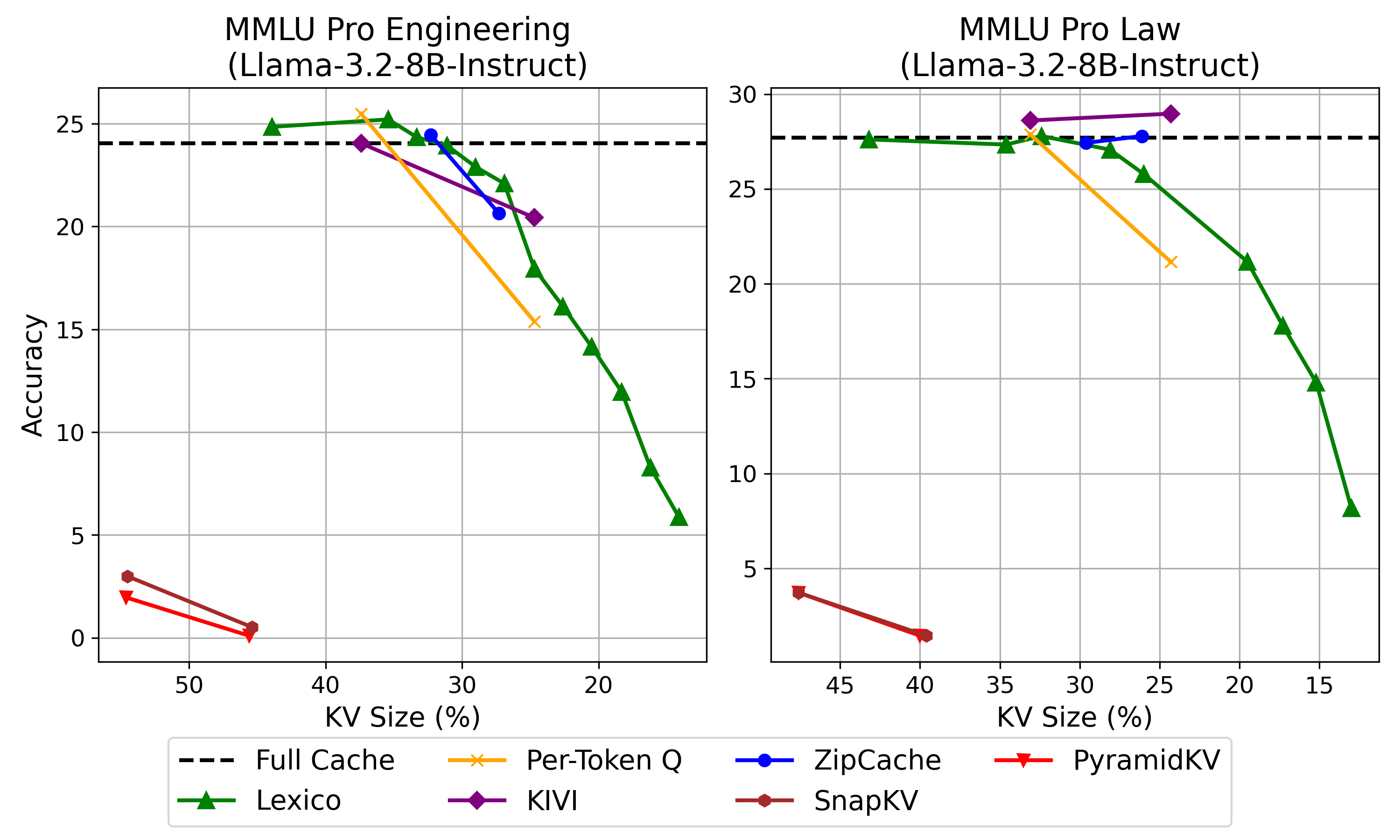}
        \caption{
            \textbf{Memory usage vs. performance of Llama-3.2-8B-Instruct with \algoname on MMLU-Pro Engineering/Law.} 
            For \algoname, we use $N = 4096$ as the dictionary size and $n_b = 128$ as the buffer size. 
            \algoname often outperforms both eviction-based methods (SnapKV, PyramidKV) and quantization-based methods (per-token quantization, KIVI, ZipCache). 
            For Law, our method slightly underperforms around 25\%, but in lower memory regimes, our method still outperforms any other baseline.
        }
        \label{fig:performance_vs_memory_usage_mmlu_pro}
    \end{minipage}
\end{figure}

\paragraph{Results across model sizes and baselines.}
We illustrate the trade-off between memory usage and performance across six different \gls{kv} cache compression methods on Llama models (1B, 3B, and 8B) in Figure~\ref{fig:performance_vs_memory_usage_gsm8k}. For all three model sizes, \algoname consistently lies on the Pareto frontier, achieving higher scores than other compression methods at similar \gls{kv} cache budget sizes. Notably, \algoname demonstrates greater robustness at smaller model scales, with larger performance gaps observed for the 1B and 3B models. 
% By controlling sparsity, \algoname can efficiently operate under a wide range of KV cache budgets. 
In the extremely low-memory regime below $20\%$, where quantization methods such as KIVI and ZipCache cannot achieve feasible cache sizes, \algoname achieves superior performance. Furthermore, while eviction-based methods (SnapKV, PyramidKV) can operate in these extremely low-memory settings, their performance lags significantly behind due to their incompatibility with \gls{gqa}, making \algoname the effective choice for stringent memory constraints.
We also evaluate \algoname on a larger model, Qwen2.5-14B-Instruct, with its weights quantized to 4 bits, comparing it against quantization methods. The results, illustrated in Figure~\ref{fig:performance_vs_memory_usage_qwen_gsm8k}, show that \algoname achieves a higher GSM8K score than KIVI under similar \gls{kv} cache budgets. Additionally, \algoname enables higher compression ratios than 2-bit quantization methods, facilitating deployment under extreme memory-constrained scenarios.

\paragraph{MMLU-Pro results.} Figure~\ref{fig:performance_vs_memory_usage_mmlu_pro} illustrates the trade-offs between memory usage and performance for \algoname on the MMLU-Pro Engineering and Law subjects using the Llama-3.1-8B-Instruct model. 
\algoname outperforms eviction-based methods like SnapKV and PyramidKV across all memory settings, though its performance is comparable to quantization-based methods such as KIVI and ZipCache.
However, in a low memory regime below 20\% cache, our method still outperforms any other baseline. This highlights that \algoname supports a wide range of compression ratios quite effectively and that our dictionary is generalizable across input distributions.

\subsection{Ablation Study}

In this subsection, we ablate the various components of \algoname and how it may influence memory complexity and task performance.

\subsubsection{Error Thresholding in Sparse Approximation} \label{sec:error_thresholding}

\algoname also supports a quality-controlled method for memory saving by allowing early termination of the sparse approximation process when a predefined error threshold is met. This approach conserves memory that would otherwise be used for marginal improvements in approximation quality. 

For the error thresholding ablation study, detailed results are provided in Table~\ref{tab:error_thresholding}. We set a maximum sparsity of 32, corresponding to the maximum number of iterations for the \gls{omp} algorithm. However, if the reconstruction error at any iteration falls below a predefined error threshold, we let the \gls{omp} terminate early, saving memory that would otherwise be used for minor approximation improvements. This approach is particularly compatible with \gls{omp}, as its greedy nature ensures that early termination yields the same results as using higher sparsity (less non-zero elements). Additionally, \gls{omp} inherently computes the residual at each iteration, allowing for continuous evaluation of the relative reconstruction error without requiring any additional computation.

\begin{table}[htbp]
    \centering
    \scriptsize  % You can try \tiny if needed, but it may affect readability
    \setlength{\tabcolsep}{4pt}  % Adjust column padding (default is 6pt)
    \caption{\textbf{Impact of error thresholding on LongBench performance and memory usage.} The table presents the performance of \algoname on the Llama-3.1-8B-Instruct model at various reconstruction error thresholds ($\delta$) for early termination of the sparse approximation algorithm. A dictionary size of $N = 1024$ and \texttt{FP16} precision for the values of the \gls{csr} tensors are used.}
    \label{tab:error_thresholding}
    \begin{tabular}{@{}lccccccccccc@{}}
        \toprule
        \textbf{Threshold ($\delta$)} & \textbf{KV Size} & \textbf{Qasper} & \textbf{QMSum} & \textbf{MultiNews} & \textbf{TREC} & \textbf{TriviaQA} & \textbf{SAMSum} & \textbf{LCC} & \textbf{RepoBench-P} & \textbf{Average} \\ 
        \midrule
        
        % Llama 3.1 8B Instruct Section
        \multicolumn{11}{c}{\textbf{Llama-3.1-8B-Instruct}} \\  % Model name spanning 11 columns
        Full Cache   & $100\%$ & $22.54$ & $24.57$ & $27.44$ & $72.5$ & $91.65$ & $43.47$ & $63.15$ & $56.76$ & $50.26$ \\
        $0.2$ & $50.6\%$ & $20.03$ & $23.65$ & $26.44$ & $72.5$ & $91.61$ & $43.47$ & $62.72$ & $56.63$ & $49.63$ \\
        $0.3$ & $41.1\%$ & $16.49$ & $23.35$ & $25.34$ & $72.5$ & $91.34$ & $43.02$ & $62.53$ & $56.65$ & $48.90$ \\
        $0.4$ & $30.9\%$ & $16.08$ & $22.91$ & $23.77$ & $69.5$ & $90.79$ & $42.70$ & $61.28$ & $54.82$ & $47.73$ \\
        $0.5$ & $22.8\%$ & $12.43$ & $21.75$ & $21.29$ & $57.5$ & $88.56$ & $41.04$ & $58.85$ & $53.19$ & $44.33$ \\
        
        \bottomrule
    \end{tabular}
\end{table}

\subsubsection{Balancing memory between buffer and sparse representation} 
We examine how balancing memory allocation between the buffer and the sparse representation affects performance as shown in Table~\ref{tab:buffer_ablation}. Fixing the total \gls{kv} cache size at 25\% of the original, we vary the memory distribution between the buffer and the sparse representation across three LongBench tasks. Qasper, MultiNews, and TREC. The results demonstrate that the ability to understand long-contexts using \algoname is not based solely on the buffer or the sparse representation. Rather, there exist optimal balance points where performance is maximized for each task.

\begin{table}[ht]
    \centering
    \small  % Adjust size as needed
    \setlength{\tabcolsep}{6pt}  % Adjust column padding
    \caption{\textbf{Balancing memory between buffer and sparse representation.} This table shows the performance of \algoname\ with the Llama-3.1-8B-Instruct model on LongBench tasks (Qasper, MultiNews, TREC) while varying the memory allocation between the buffer and the sparse representation, with the total \gls{kv} cache size fixed at 25\% of the original size.}
    \label{tab:buffer_ablation}
    \begin{tabular}{@{} c c c c c c c c c @{}}
        \toprule
        \multicolumn{3}{c}{\textbf{Qasper}} & \multicolumn{3}{c}{\textbf{MultiNews}} & \multicolumn{3}{c}{\textbf{TREC}} \\  
        \cmidrule(r){1-3} \cmidrule(r){4-6} \cmidrule(r){7-9}
        $\mathbf{s}$ & $\mathbf{n_b}$ & \textbf{F1 Score} & $\mathbf{s}$ & $\mathbf{n_b}$ & \textbf{ROUGE-L} & $\mathbf{s}$ & $\mathbf{n_b}$ & \textbf{Accuracy} \\ 
        \midrule
        $1$   & $862$ & $6.38$   & $1$   & $503$ & $17.20$ & $1$   & $1232$ & $58.5$ \\
        $4$   & $724$ & $8.36$   & $4$   & $423$ & $20.21$ & $4$   & $1035$ & $63.5$ \\
        $8$   & $517$ & $14.58$  & $8$   & $302$ & $21.27$ & $\mathbf{8}$   & $\mathbf{739}$  & $\mathbf{65.0}$ \\
        $\mathbf{12}$  & $\mathbf{278}$ & $\mathbf{17.84}$  & $\mathbf{12}$ & $\mathbf{163}$ & $\mathbf{22.81}$ & $12$  & $398$  & $63.5$ \\
        $16$  & $0$   & $8.27$   & $16$  & $0$   & $10.70$ & $16$  & $0$    & $54.5$ \\
        \bottomrule
    \end{tabular}
\end{table}

\subsubsection{Performance without buffer} 
\label{sec:wo_buffer}
\algoname incorporates a buffer that retains the most recent $n_b$ tokens in full precision, similarly to prior studies that find that this buffer is crucial to maintain performance. Our finding for \algoname also aligns closely with this observation.

\begin{figure}[htb]
    \centering
    \includegraphics[width=\linewidth]{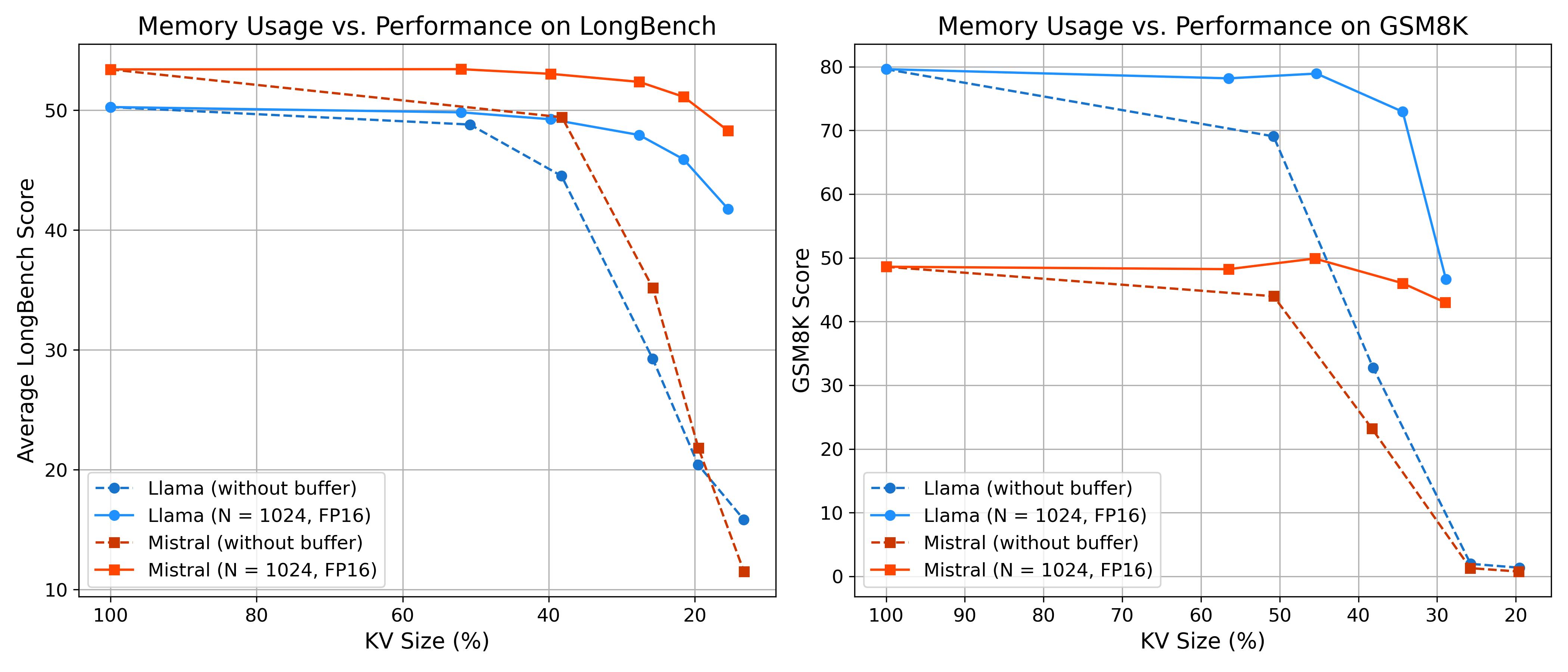}
    \caption{
        \textbf{Memory usage vs. performance of \algoname with and without buffer on LongBench and GSM8K.}
        The figure illustrates the impact of removing the buffer on the performance of \algoname when evaluated on the Llama-3.1-8B-Instruct and Mistral-7B-Instruct models for LongBench (left) and GSM8K (right) tasks. Solid lines represent configurations with a buffer, while dashed lines represent configurations without a buffer. We use a dictionary size of $N = 1024$ and \texttt{FP16} precision for the values of \gls{csr} tensors to vary sparsity and explore a wide range of \gls{kv} sizes. 
    }
    \label{fig:performance_vs_memory_usage}
\end{figure}

To evaluate the impact of the buffer, we first conduct experiments with varying sparsity without the buffer, with the results shown by the dashed lines in Figure~\ref{fig:performance_vs_memory_usage}. The comparison shows that removing the buffer results in a more pronounced decline in performance, especially at lower \gls{kv} sizes. 
Full numerical results are shown in Table~\ref{tab:longbench_wo_buffer} and Table~\ref{tab:gsm8k_wo_buffer} in Appendix~\ref{sec:effect_of_buffer}.

\subsubsection{Adaptive Dictionary Learning} \label{sec:adaptive}

While our universal dictionaries demonstrate strong performance, we explore an adaptive learning method to better incorporate input-specific context. This adaptive approach improves performance by adding new dictionary atoms during generation when the predefined reconstruction error threshold is not met. These atoms, tailored to the input prompt, improve performance, but cannot be shared across batches, which requires them to be included in the \gls{kv} size calculation. Although this approach boosts accuracy, it increases memory usage, limiting its ability to achieve low-memory regimes.

Though we observe some degree of universality in our dictionaries, as shown in Table~\ref{tab:dictionary_comparison}, their performance is particularly strong on WikiText-103, the dataset they were trained on. To better incorporate input context information, we propose an extension that adaptively learns the dictionary during generation. 

In this framework, we begin with a pre-trained universal dictionary as the initial dictionary. If, during the generation process, the sparse approximation fails to meet the predefined relative reconstruction error threshold, the problematic uncompressed key or value vector is normalized and added to the dictionary. The sparse representation of this vector is then stored with a sparsity of $s=1$, where its index corresponds to the newly added atom and its value is the $\ell_2$ norm of the uncompressed vector. The updated dictionary is subsequently used for further sparse approximations during the generation task. In this way, the adaptive learning framework incrementally refines the dictionaries, tailoring them to the specific generative task and enhancing overall performance at the cost of additional memory usage.

In our experiment, we initialize with a dictionary of size 1024 derived from WikiText-103; we then incorporate up to 1024 additional atoms during inference, resulting in a total dictionary size of $N = 2048$. The maximum sparsity of $s = 16$ is used, with a buffer size of $n_b = 128$, and \texttt{FP16} precision for the values of the \gls{csr} tensors. Results of this experiment are presented in Table~\ref{tab:adaptive_gsm}.
For both the Llama-3.1-8B-Instruct and Mistral-7B-Instruct models, the best GSM8K scores were observed when the relative reconstruction error threshold was set to $\delta = 0.3$. Under this setting, the adaptive \algoname achieved improvements of $4.5\%p$ and $2.1\%p$ in the GSM8K score compared to the baseline \algoname for the Llama and Mistral models, respectively. The baseline \algoname uses a static initial dictionary of size 1024 without any adaptation. However, these improvements come at the cost of increased \gls{kv} cache sizes of $9.1\%$ for Llama and $6.9\%$ for Mistral.

% \textcolor{red}{TODO: analysis sentences. How do we know it did better? also list universal dictionary accuracy as baseline?}

\begin{table}[htbp]
    \centering
    \small  % Adjust size as needed
    \setlength{\tabcolsep}{4pt}  % Adjust column padding
    \caption{\textbf{GSM8K performance of adaptive \algoname .} The table shows the GSM8K performance and \gls{kv} cache sizes of adaptive \algoname on the Llama-3.1-8B-Instruct and Mistral-7B-Instruct-v0.3 models at varying reconstruction error thresholds ($\delta$). A universal dictionary of size 1024 is used, with up to 1024 additional atoms added during generation. The maximum sparsity of $s = 16$, buffer size of $n_b = 128$, and \texttt{FP16} precision for \gls{csr} tensor values are applied.}
    \label{tab:adaptive_gsm}
    \begin{tabular}{@{}lccccc@{}}
        \toprule
        \textbf{Threshold ($\delta$)} & \multicolumn{2}{c}{\textbf{Llama-3.1-8B-Instruct}} & \multicolumn{2}{c}{\textbf{Mistral-7B-Instruct-v0.3}} \\ 
        \cmidrule(lr){2-3} \cmidrule(lr){4-5}
        & \textbf{\gls{kv} Size} & \textbf{GSM8K Score} & \textbf{\gls{kv} Size} & \textbf{GSM8K Score} \\ 
        \midrule
        
        Full Cache     & $100\%$   & $79.61$ & $100\%$   & $48.60$ \\
        w/o Adaptation & $34.4\%$  & $72.93$ & $34.4\%$  & $46.02$ \\
        $0.25$         & n/a       & n/a     & $42.1\%$  & $48.07$ \\
        $0.30$         & $43.5\%$  & $\mathbf{77.41}$ & $41.3\%$  & $\mathbf{48.14}$ \\
        $0.35$         & $42.0\%$  & $76.80$ & $39.8\%$  & $47.76$ \\
        
        \bottomrule
    \end{tabular}
\end{table}

\subsection{Latency Analysis}
\label{sec:latency}

In this section, we present latency measurements of the forward pass and \gls{omp} portion of \algoname during decoding stage in Table~\ref{tab:latency}.
We run simple generation tests on a 1000 token input to Llama-3.1-8B-Instruct model and generate up to 250 tokens to measure and aggregate latency metrics.
We compare both dictionary sizes $N=1024$ and $4096$, which primarily affects \gls{omp} computation time.
We set the sparsity level to $s=24$, and process OMP in batches of $n_a = 8$.

Although we list the forward pass and \gls{omp} separately, the processes are implemented to run in parallel such that the one generation step takes the maximum of the two durations plus some overhead. 
However, with parallelization, there exists a time versus space complexity tradeoff, since running OMP also consumes GPU memory.

Higher latency of \algoname may be a limitation for latency-critical use cases. 
However, our primary focus is addressing highly memory-constrained scenarios. Such scenarios are increasingly critical for real-world LLM deployments, where even a batch size of one can exceed the capacity of a single GPU. By prioritizing memory efficiency, \algoname enables feasible deployment in contexts where other methods may encounter out-of-memory errors, offering a crucial contribution to memory-limited settings.

\begin{table}[tb]
    \centering
    \setlength{\tabcolsep}{4pt}  % Adjust column padding
    \caption{\textbf{Latency measurements.} The following latencies measure the total time it takes for the respective computation to process when generating one new token. We use Llama-3.1-8B-Instruct and sum up the time each operation takes in total across all 32 layers. Latencies when dictionary sizes are 1024 and 4096 are measured. Detailed settings are described in Section~\ref{sec:latency}.}
    \begin{subtable}{0.8\textwidth}
        \centering
        \begin{tabular}{@{}lcc@{}}
            \toprule
            \textbf{Computation Type} & \multicolumn{2}{c}{\textbf{Latency} (per token)} \\ 
            \cmidrule(lr){2-3}
            - & \textbf{$N=1024$} & \textbf{$N=4096$} \\ 
            \midrule
            Standard forward pass $(\vq \mK^\top)$  & $48.39$ ms & -- \\
            \algoname: forward pass using $\vq ( \mK_{\text{csr}} \mD_k^\top)^\top$  & $55.56$ ms & $56.35$ ms \\
            % \algoname: forward pass using $(\vq \mD_k) \mK_{\text{csr}}^\top$  & $-$ ms & $-$ ms \\
            \algoname: sparse approximation via \gls{omp}  & $26.57$ ms & $40.58$ ms \\
            \bottomrule
        \end{tabular}
    \end{subtable}
    \label{tab:latency}
\end{table}

\section{Conclusion}

In conclusion, our proposed method, \algoname, offers a novel approach to compressing KV cache for transformers by leveraging low-dimensional structures and sparse dictionary learning. 
Through this method, we demonstrate that substantial redundancy exists among key cache across various inputs, allowing us to compress the KV cache efficiently while maintaining near-lossless performance. 
Furthermore, \algoname enables compression rates that surpass traditional quantization techniques, offering fine-grained and wide control over memory usage. Importantly, our universal dictionary is both compact and scalable, making it applicable across tasks and user inputs without increasing memory demands. 
This approach provides strong memory savings, particularly for long-context tasks, by alleviating the memory bottlenecks associated with KV cache storage without dropping any previous tokens. 

Future research directions based on our work include optimizing CSR tensors through customized quantizations and improving latency tradeoffs that occur due to the use of OMP during prefilling and decoding.
It would also be interesting to apply ``soft-eviction strategies'' for sparse tensors in which sparsity level $s$ is determined or dropped later on based on the estimated importance of the token. A dynamic allocation of sparsity can further improve our compression method.

\bibliography{iclr2025}
\bibliographystyle{iclr2025}

\newpage
\appendix

\section*{Appendix}

\section{Implementation Details}
\label{sec:omp_implementation_details}

Algorithm~\ref{alg:omp_simple} illustrates a naive implementation of \gls{omp} for understanding. In \algoname , we adopt the implementation of \gls{omp} v0 proposed by \citep{zhu2020efficient}, which minimizes computational complexity using efficient inverse Cholesky factorization. Additionally, we integrate methods from \citep{lubonja2024efficient} for batched GPU execution and extend the implementation to handle multiple dictionaries in parallel. Algorithm~\ref{alg:lexico_algorithm} presents the pseudocode for \algoname during the prefilling and decoding stages. 

\begin{algorithm}
\caption{\gls{omp}}\label{alg:omp_simple}
\begin{algorithmic}[1]
    \Require Signal $\vk \in \R^m$, dictionary $\mD \in \R^{m \times N}$, sparsity $s$
    \Ensure Sparse representation $\vy \in \R^n$
    \State Initialize $\vr^{(0)} \gets \vk$, $\sI^{(0)} \gets \emptyset$, $\vy^{(0)} \gets \mathbf{0}$
    \For{$i = 1$ to $s$}
        \State $n^{(i)} \gets \arg\max_{1 \leq n \leq N} \left\{ \left| \left( \mD^\top \left( \vk - \mD \vy^{(i)} \right) \right)_n \right| \right\}$
        \State $\sI^{(i)} \gets \sI^{(i-1)} \cup \left\{ n^{(i)} \right\}$
        \State $\vy^{(i+1)} \gets \arg\min_{\vy \in \sR^N} \left\{ \left\| \vk - \mD \vy \right\|_2, \text{Supp} \left( \vy \right) \subset \sI^{(i)} \right\}$
    \EndFor
    \State \Return $\vy$
\end{algorithmic}
\end{algorithm}

\new{\begin{algorithm}
\caption{\new{Prefilling and decoding with \algoname}}\label{alg:lexico_algorithm}
\begin{algorithmic}[1]
    \State \textbf{Parameter:} sparsity $s$, buffer length $n_b$, approximation length $n_a$
    \Statex
    % \State \textbf{Procedure:} Prefill
    \Procedure{Prefilling}{}
        \State \textbf{Input:} $\mX \in \R^{l_{\text{seq}} \times d}$
        \State $\mK \gets \mX \mW_k$, $\mV \gets \mX \mW_v$
        \State $\mK_{\text{csr}} \gets \text{OMP} \left ( \mK \left [ : l_{\text{seq}} - n_b \right ], \mD_k, s \right )$
        \State $\mV_{\text{csr}} \gets \text{OMP} \left ( \mV \left [ : l_{\text{seq}} - n_b \right ], \mD_v, s \right )$
        \State $\mK_{\text{buffer}} \gets \mK \left [l_{\text{seq}} - n_b : \right ]$, $\mV_{\text{buffer}} \gets \mV \left [l_{\text{seq}} - n_b : \right ]$
        \State $\text{KV cache} \gets \mK_{\text{csr}}, \mK_{\text{buffer}}, \mV_{\text{csr}}, \mV_{\text{buffer}}$
        \State \Return $\mK$, $\mV$ 
    \EndProcedure
    \Statex
    \Procedure{Decoding}{}
        \State \textbf{Input:} $\text{KV cache}$, $\vx_t \in \R^{1 \times d}$
        \State $\vq_t \gets \vx_t \mW_q$, $\vk_t \gets \vx_t \mW_k$, $\vv_t \gets \vx_t \mW_v$
        \State $\mK_{\text{csr}}, \mK_{\text{buffer}}, \mV_{\text{csr}}, \mV_{\text{buffer}} \gets \text{KV cache}$
        \State $\mK_{\text{buffer}} \gets \text{Concat} \left ( \left [ \mK_{\text{buffer}}, \vk_t \right ], \text{dim} = \text{token} \right )$
        \State $\mV_{\text{buffer}} \gets \text{Concat} \left ( \left [ \mV_{\text{buffer}}, \vv_t \right ], \text{dim} = \text{token} \right )$
        \State $\va_t \gets \text{Concat} \left ( \left [ \vq_t \mD_k \mK_{\text{csr}}, \vq_t \mK_{\text{buffer}} \right ], \text{dim} = \text{token} \right )$
        \State $\va_t \gets \text{Softmax} \left ( \va_t \right )$
        \State $\mV \gets \text{Concat} \left ( \left [ \mD_v \mV_{\text{csr}}, \mV_{\text{buffer}} \right ], \text{dim} = \text{token} \right )$
        \State $\vo_t \gets \va_t \mV$
        \If{$\text{len} \left ( \mK_{\text{buffer}} \right ) > n_b$}
            \State ${\mK'}_{\text{csr}} \gets \text{OMP} \left (\mK_{\text{buffer}} \left [ : n_a \right ], \mD_k, s \right )$
            \State ${\mV'}_{\text{csr}} \gets \text{OMP} \left (\mV_{\text{buffer}} \left [ : n_a \right ], \mD_v, s \right )$
            \State $\mK_{\text{csr}} \gets \text{Concat} \left ( \left [ \mK_{\text{csr}}, {\mK'}_{\text{csr}} \right ], \text{dim} = \text{token} \right )$
            \State $\mV_{\text{csr}} \gets \text{Concat} \left ( \left [ \mV_{\text{csr}}, {\mV'}_{\text{csr}} \right ], \text{dim} = \text{token} \right )$
            \State $\mK_{\text{buffer}} \gets \mK_{\text{buffer}} \left [ n_a : \right ]$, $\mV_{\text{buffer}} \gets \mV_{\text{buffer}} \left [ n_a : \right ]$
        \EndIf
        \State $\text{KV cache} \gets \mK_{\text{csr}}, \mK_{\text{buffer}}, \mV_{\text{csr}}, \mV_{\text{buffer}}$
        \State \Return $\vo_t$ 
    \EndProcedure

\end{algorithmic}
\end{algorithm}}

\section{LongBench Task Statistics} \label{sec:longbench_task_statistics}

\begin{table}[htbp]
    \centering  % You can try \tiny if needed, but it may affect readability
    \small
    \setlength{\tabcolsep}{4pt}  % Adjust column padding (default is 6pt)
    \caption{\textbf{Details of LongBench tasks used in experiments.}}
    \label{tab:longbench_task_statistics}
    \begin{tabular}{@{}lcccc@{}}
        \toprule
        \textbf{Task} & \textbf{Task Type} & \textbf{Evaluation Metric} & \textbf{Average Length} & \textbf{\# of Samples} \\
        \midrule
        Qasper       & Single-doc QA          & F1           & 3619  & 200  \\
        QMSum        & Summarization          & ROUGE-L      & 10614 & 200  \\
        MultiNews    & Summarization          & ROUGE-L      & 2113  & 200  \\
        TREC         & Few-shot information retrieval & Accuracy     & 5177   & 200  \\
        TriviaQA     & Few-shot reading comprehension & F1           & 8209 & 200  \\
        SAMSum       & Few-shot dialogue summarization & ROUGE-L      & 6258  & 200 \\
        LCC          & Code completion        & Edit Similarity    & 1235  & 500  \\
        RepoBench-P  & Code completion        & Edit Similarity    & 4206  & 500  \\
        \bottomrule
    \end{tabular}
\end{table}

\section{Ablation Study: Experimental Results}

\subsection{Performance without Buffer} \label{sec:effect_of_buffer}

In this section, we assess the effect of the buffer by comparing the results with and without its use. We illustrated the result in Figure~\ref{fig:performance_vs_memory_usage} in Section~\ref{sec:wo_buffer}. The quantitative results for LongBench and GSM8K are presented in Table~\ref{tab:without_buffer_longbench} and Table~\ref{tab:without_buffer_gsm}, respectively.

\begin{table}[htbp]
    \centering
    \scriptsize  % You can try \tiny if needed, but it may affect readability
    \setlength{\tabcolsep}{4pt}  % Adjust column padding (default is 6pt)
    \caption{\textbf{LongBench performance without buffer.} This table shows the impact of removing the buffer of \algoname on the performance of the Llama-3.1-8B-Instruct and Mistral-7B-Instruct models at varying sparsity levels. A dictionary size of $N = 1024$ and \texttt{FP16} precision for the values of the \gls{csr} tensors are used.}
    \label{tab:without_buffer_longbench}
    \begin{tabular}{@{}lccccccccccc@{}}
        \toprule
        \textbf{Sparsity} & \textbf{KV Size} & \textbf{Qasper} & \textbf{QMSum} & \textbf{MultiNews} & \textbf{TREC} & \textbf{TriviaQA} & \textbf{SAMSum} & \textbf{LCC} & \textbf{RepoBench-P} & \textbf{Average} \\ 
        \midrule
        
        % Llama 3.1 8B Instruct Section
        \multicolumn{11}{c}{\textbf{Llama-3.1-8B-Instruct}} \\  % Model name spanning 11 columns
        Full Cache    & $100\%$ & $13.10$ & $23.46$ & $26.94$ & $72.5$ & $91.65$ & $43.47$ & $63.15$ & $56.76$ & $48.88$ \\
        $s=32$ & $50.8\%$ & $14.87$ & $26.51$ & $26.57$ & $71.5$ & $92.48$ & $42.88$ & $61.54$ & $54.04$ & $48.80$ \\
        $s=24$ & $38.2\%$ & $13.37$ & $25.02$ & $22.54$ & $65.0$ & $91.75$ & $39.71$ & $52.21$ & $46.48$ & $44.51$ \\
        $s=16$ & $25.8\%$ & $8.27$ & $13.74$ & $10.70$ & $54.5$ & $77.51$ & $20.45$ & $26.53$ & $22.46$ & $29.27$ \\
        $s=12$ & $19.5\%$ & $6.31$ & $10.15$ & $5.66$ & $39.0$ & $53.70$ & $6.83$ & $22.18$ & $19.46$ & $20.41$ \\
        $s=8$ & $13.3\%$ & $2.74$ & $8.05$ & $4.17$ & $36.5$ & $34.45$ & $4.27$ & $18.24$ & $18.32$ & $15.84$ \\
        
        \midrule
        
        % Mistral 7B Instruct Section
        \multicolumn{11}{c}{\textbf{Mistral-7B-Instruct-v0.3}} \\  % Model name spanning 11 columns
        Full Cache    & $100\%$ & $41.58$ & $25.69$ & $27.76$ & $76.0$ & $88.59$ & $47.58$ & $59.37$ & $60.60$ & $53.40$ \\
        $s=32$ & $50.8\%$ & $40.27$ & $25.21$ & $27.53$ & $76.5$ & $89.01$ & $45.77$ & $58.64$ & $59.07$ & $52.75$ \\
        $s=24$ & $38.2\%$ & $37.46$ & $24.41$ & $27.34$ & $75.5$ & $88.66$ & $43.87$ & $48.55$ & $49.50$ & $49.41$ \\
        $s=16$ & $25.8\%$ & $25.57$ & $18.49$ & $15.19$ & $71.5$ & $81.91$ & $27.90$ & $19.39$ & $21.45$ & $35.18$ \\
        $s=12$ & $19.5\%$ & $18.59$ & $13.11$ & $5.95$ & $58.0$ & $50.13$ & $2.86$ & $13.38$ & $12.60$ & $21.83$ \\
        $s=8$ & $13.3\%$ & $10.32$ & $6.98$ & $2.67$ & $31.5$ & $20.01$ & $2.27$ & $10.18$ & $8.11$ & $11.51$ \\
        \bottomrule
    \end{tabular}
    \label{tab:longbench_wo_buffer}
\end{table}

\begin{table}[htbp]
    \centering
    \small  % Adjust size as needed
    \setlength{\tabcolsep}{4pt}  % Adjust column padding
    \caption{\textbf{GSM8K performance without buffer.} This table shows the impact of removing the buffer of \algoname on the performance of the Llama-3.1-8B-Instruct and Mistral-7B-Instruct models at varying sparsity levels. A dictionary size of $N = 1024$ and \texttt{FP16} precision for the values of the \gls{csr} tensors are used.}
    \label{tab:without_buffer_gsm}
    \begin{tabular}{@{}lcccc@{}}
        \toprule
        \textbf{Sparsity} & \textbf{KV Size} & \textbf{Llama-3.1-8B-Instruct} & \textbf{Mistral-7B-Instruct-v0.3} \\ 
        \midrule
        
        Full Cache  & $100\%$  & $79.61$ & $48.60$ \\
        $s=32$      & $50.8\%$ & $69.07$ & $43.97$ \\
        $s=24$      & $38.2\%$ & $32.75$ & $23.20$ \\
        $s=16$      & $25.8\%$ & $1.97$  & $1.29$ \\
        $s=12$      & $19.6\%$ & $1.36$  & $0.76$ \\
        
        \bottomrule
    \end{tabular}
    \label{tab:gsm8k_wo_buffer}
\end{table}

\section{Qualitative Results}
\subsection{GSM8K Example Generation}
\label{app:gsm8k_example}

\new{The boxes highlight the correct intermediate answers that must be derived. The highlighted colors in red show incorrect reasoning and answers. We provide the original problem and solution below, along with outputs from Llama-3.1-8B-Instruct with full KV cache (FP16), \algoname (sparsity $s=14$), and KIVI-2.}

\begin{framed}
\colorbox{lightgray}{\parbox{\linewidth}{\textbf{Original Problem \& Solution}}}

\textbf{Problem:} The Kennel house keeps 3 German Shepherds and 2 Bulldogs. If a German Shepherd consumes 5 kilograms of dog food and a bulldog consumes 3 kilograms of dog food per day, how many kilograms of dog food will they need in a week?

\textbf{Solution:}
The German Shepherds consume $3 \times 5 = \boxed{15}$ kilograms of dog food per day. \\
The Bulldogs consume $2 \times 3 = \boxed{6}$ kilograms of dog food per day. \\
The dogs consume a total of $15 + 6 = \boxed{21}$ kilograms of dog food per day. \\
Therefore, they need $21 \times 7 = \boxed{147}$ kilograms of dog food in a week.

\textbf{Final Answer:} 147
\end{framed}

\begin{framed}
\colorbox{lavender}{\parbox{\linewidth}{\textbf{Llama-3.1-8B-Instruct with Full Cache}}}

First, find the total amount of food the German Shepherds consume in a day: $3$ dogs $\times$ $5$ kg/dog $= 15$ kg. \\
Then find the total amount of food the Bulldogs consume in a day: $2$ dogs $\times$ $3$ kg/dog $= 6$ kg. \\
Add the two amounts together to find the total amount of food the dogs consume in a day: $15$ kg $+ 6$ kg $= 21$ kg. \\
Multiply that number by 7 to find the total amount of food the dogs consume in a week: $21$ kg/day $\times 7$ days $= 147$ kg.
\end{framed}

\begin{framed}
\colorbox{lavender}{\parbox{\linewidth}{\textbf{Llama-3.1-8B-Instruct with \algoname ($s=14$)}}}

The kennel has $3$ German Shepherds and $2$ Bulldogs, so they have $3 + 2 = 5$ dogs. \\
The German Shepherds consume $5 \times 3 = 15$ kilograms of dog food per day. \\
The Bulldogs consume $2 \times 3 = 6$ kilograms of dog food per day. \\
So, the kennel needs $15 + 6 = 21$ kilograms of dog food per day. \\
In a week, they will need $21 \times 7 = 147$ kilograms of dog food. \\
The answer is 147.
\end{framed}

\begin{framed}
\colorbox{lavender}{\parbox{\linewidth}{\textbf{Llama-3.1-8B-Instruct with KIVI-2}}}

The Kennel house has $3$ German Shepherds and $2$ Bulldogs, so they have a total of $3 + 2 = \boxed{5}$ dogs. \\
Each German Shepherd consumes $5$ kilograms of dog food per day, so the total amount of dog food consumed by the German Shepherds is \textcolor{red}{$5 \times 5 = \boxed{25}$} kilograms per day. \\
Each Bulldog consumes $3$ kilograms of dog food per day, so the total amount of dog food consumed by the Bulldogs is $2 \times 3 = \boxed{6}$ kilograms per day. \\
The total amount of dog food consumed per day is \textcolor{red}{$25 + 6 = \boxed{31}$} kilograms. \\
The Kennel house will need \textcolor{red}{$31 \times 7 = \boxed{217}$} kilograms of dog food in a week.
\end{framed}

\end{document}